\documentclass[letterpaper,journal]{IEEEtran}
\usepackage{amsmath,amsfonts,amssymb}
\usepackage{algorithmic}
\usepackage{algorithm}
\usepackage{array}
\usepackage{booktabs}
\usepackage{multirow}
\usepackage[caption=false,font=normalsize,labelfont=sf,textfont=sf]{subfig}
\usepackage{textcomp}
\usepackage{stfloats}
\usepackage{url}
\usepackage{verbatim}
\usepackage{graphicx}
\usepackage{cite}
\usepackage{xcolor}
\usepackage{xspace}
\usepackage[hidelinks]{hyperref}
\newcommand{\ours}{Sat2City v2\xspace}

\newcommand{\cmark}{\textcolor{green!60!black}{\ensuremath{\checkmark}}}
\newcommand{\xmark}{\textcolor{red}{\ensuremath{\times}}}
\newcommand{\pmark}{\textcolor{orange!85!black}{\ensuremath{\triangle}}}

\begin{document}
\bstctlcite{IEEEtranBSTCTL:nodash}

\title{\ours: Native 3D City Asset Generation from a Single Satellite Image}

\author{Tongyan Hua\textsuperscript{\dag}, Dongli Wu\textsuperscript{\dag}, Jinjing Zhu, Yinrui Ren, Zhongcheng Hong, Ying-Cong Chen, \\
Hui Xiong,~\IEEEmembership{Fellow, IEEE}, and Wufan Zhao%
\thanks{Tongyan Hua, Dongli Wu, Yinrui Ren, Ying-Cong Chen, Hui Xiong, and Wufan Zhao are with The Hong Kong University of Science and Technology (Guangzhou), Guangzhou, China (e-mail: thua388@connect.hkust-gz.edu.cn; dwu022@connect.hkust-gz.edu.cn; conorren001@gmail.com; yingcongchen@ust.hk; xionghui@ust.hk; wufanzhao@hkust-gz.edu.cn).}%
\thanks{Jinjing Zhu is with The Chinese University of Hong Kong, Hong Kong SAR, China (e-mail: jinjingzhu.mail@gmail.com).}%
\thanks{Zhongcheng Hong completed this work while with The Hong Kong University of Science and Technology (Guangzhou), Guangzhou, China, and is now with Auckland University of Technology, Auckland, New Zealand (e-mail: zhongcheng.hong@autuni.ac.nz).}%
\thanks{Ying-Cong Chen and Hui Xiong are also with the Department of Computer Science and Engineering, The Hong Kong University of Science and Technology, Hong Kong SAR, China.}%
\thanks{\textsuperscript{\dag}Tongyan Hua and Dongli Wu contributed equally to this work. Corresponding author: Wufan Zhao.}%
\thanks{Project page: \url{https://ai4city-hkust.github.io/Sat2City-v2/}.}}

%\markboth{IEEE Transactions on Pattern Analysis and Machine Intelligence}%
%{\ours}

\IEEEaftertitletext{%
\vspace{-0.4\baselineskip}
\begin{center}
  \centering
  \includegraphics[width=0.99\textwidth]{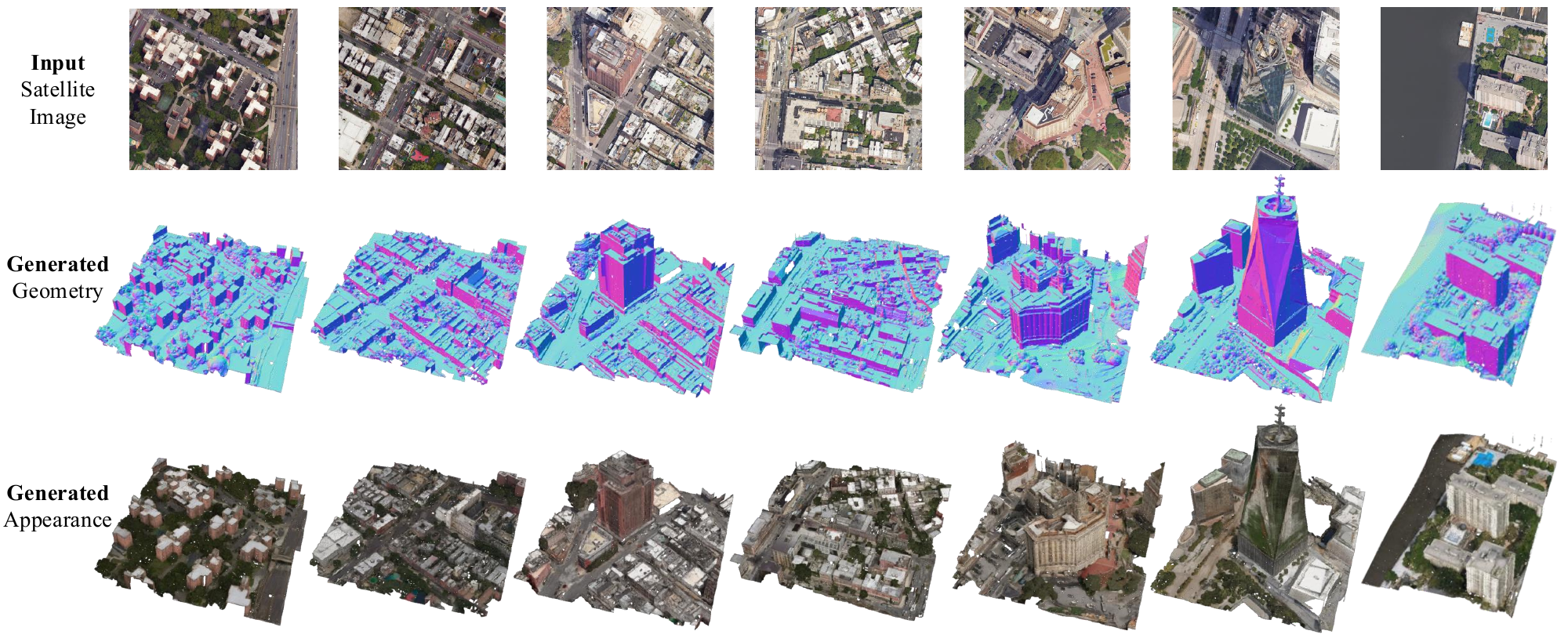}
  \vspace{0.25\baselineskip}
  \refstepcounter{figure}%
  \label{fig:teaser}
  {\footnotesize Fig.~\thefigure. Given a single satellite image, \ours generates explicit city-scale geometry and a satellite-conditioned textured output. The examples cover diverse urban layouts, from dense street blocks to landmark buildings and waterfront scenes.\par}
\end{center}
\vspace{-0.2\baselineskip}
}

\maketitle

\begin{abstract}
Generating explicit textured 3D city assets from a single satellite image is important for urban simulation and digital twins.
Most prior methods, however, learn 3D proxies optimized for rendering street-view images or videos over prescribed viewpoints and trajectories, rather than producing explicit 3D assets.
Our previous framework, Sat2City, took a first step toward this goal with task-specific cascaded sparse-voxel latent diffusion conditioned on satellite-derived height maps.
However, it relied on synthetic data and did not condition appearance on the satellite input. 
To address these limitations, we present \ours, a framework for generating explicit textured 3D city assets directly from real satellite images.
We first construct a real-world dataset of 16,027 geographically matched but weakly aligned satellite-image--textured-mesh pairs, comprising 14,651 training pairs and 1,376 held-out test pairs.
Building on this dataset, \ours uses a native structured-latent 3D prior pretrained on curated assets to anchor geospatial adaptation to its asset manifold. 
We empirically demonstrate that weakly aligned satellite images and regional textured 3D meshes can support asset-level conditional generation: 
learned attention lets satellite tokens steer geometry generation without calibrated pixel-to-surface correspondence, while the generated geometry anchors satellite-guided material synthesis.
\ours ranks first among the evaluated baselines on every reported metric across geometry accuracy, generative mesh quality, and satellite-to-asset feature alignment.
\end{abstract}


\suppressfloats[t]
\begin{figure}[t]
\raggedright
\includegraphics[width=\columnwidth]{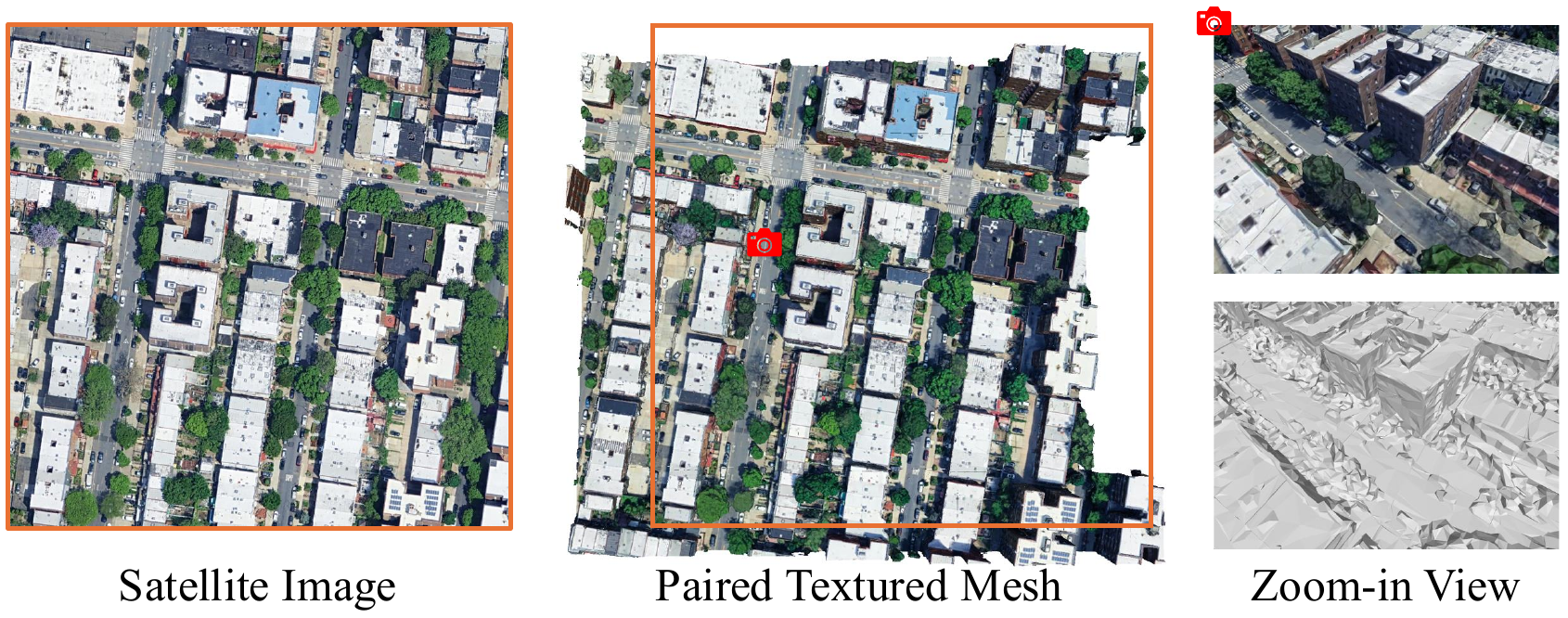}
\caption{\textbf{Geographically matched satellite-image--textured-mesh pairs in the Sat2City v2 dataset.} Each pair comprises a satellite image and a textured mesh covering the same geographic crop. Because calibrated pixel-to-surface correspondence is unavailable, the pair is treated as weakly aligned. Enlarged views highlight the geometry and appearance information available for asset-level supervision.}
\label{fig:dataset_challenge}
\end{figure}

\section{Introduction}
\IEEEPARstart{G}{enerating} 3D urban scenes~\cite{city_gen-li2024sat2scene,city_gen-lin2023infinicity,city_gen-xie2024citydreamer,xie2025generative} has attracted increasing attention in recent years, driven by applications in gaming~\cite{sce-xu2024sketch2scene}, urban planning~\cite{vid_img-lu2024urban,city_agent-yang2024procedural}, autonomous simulation~\cite{lu2024infinicube,lin2022urbanscene3d}, and digital twin systems~\cite{city_agent-shang2024urbanworld,city_agent-zhang2024cityx}.
Much prior work models cities as 3D-aware image-generation problems and primarily targets novel street-view images or roaming videos from satellite or cross-view inputs~\cite{vid_img-li2021sat2vid,vid_img-lu2020geometry,qian2023sat2density,vid_img-shi2022geometry,vid_img-xu2024geospecificviewgeneration,vid_img-li2024crossviewdiff,vid_img-deng2024streetscapes,vid_img-lu2024urban,vid_img-yang2023urbangiraffe,vid_img-li2024syntheocc}.
These methods are valuable for view synthesis, but the learned 3D representations are usually optimized as rendering proxies.
They often provide limited control over reusable scene contents, and the extracted geometry can be noisy, incomplete, or difficult to edit.
In this work, we focus on a more asset-centric problem: generating an explicit textured 3D city mesh from a single satellite image.
The satellite image is a compact and widely available control signal that encodes road layout, building footprints, vegetation, water, roofs, and dominant appearance patterns.
Turning this signal into a reusable 3D asset, however, requires both reliable surface geometry and a coherent textured output conditioned on the satellite image.
Fig.~\ref{fig:teaser} illustrates the target output: a generated 3D city asset whose geometry and texture are both conditioned on the same overhead observation.

Recent urban generation methods have introduced stronger geospatial priors to improve control.
InfiniCity~\cite{city_gen-lin2023infinicity} lifts satellite-derived height fields into volumetric features and learns NeRF-based urban generators.
CityDreamer~\cite{city_gen-xie2024citydreamer} and its generative Gaussian variants~\cite{xie2025generative} further decompose urban scenes by semantic categories and improve visual fidelity.
Sat2Density~\cite{qian2023sat2density} and Sat2Density++~\cite{qian2026sat2densitypp} directly use satellite imagery to condition radiance fields for street-view rendering.
More recently, Sat3DGen~\cite{qiansat3dgen} improves satellite-to-street 3D proxies by adding geometry-oriented constraints and a perspective-view training strategy.
Nevertheless, these approaches still learn implicit or proxy representations mainly under 2D supervision from satellite-ground image pairs.
Their geometry is judged primarily through rendered views or depth maps, not through explicit asset quality.
Consequently, even when the rendered images are photorealistic, the resulting 3D scenes can be distorted, fragmented, or unsuitable for downstream asset workflows.

Sat2City v1~\cite{Hua_2025_ICCV} took a different path by learning explicit 3D geometry and appearance under direct 3D supervision.
It introduced a task-specific cascaded latent diffusion framework over sparse voxel grids, with a Re-Hash VAE bottleneck for stable appearance optimization and inverse sampling for smooth color supervision.
Unlike rendering-oriented methods, this 3D-native formulation directly generated explicit city-scale assets from height-map observations.
However, two limitations become critical when moving to real satellite imagery and reconstructed meshes.
First, its task-specific latent representation and generator were learned from clean synthetic city assets paired with simulated height maps, leaving the framework untested on real satellite imagery and noisy reconstructed meshes.
Second, height-only conditioning provided no appearance evidence, leaving the generated appearance driven mainly by geometry and learned priors rather than by satellite cues.

Addressing these limitations first requires a real-world dataset that associates each satellite image with a textured 3D city mesh from the same geographic region.
Existing satellite-to-ground benchmarks, such as CVUSA~\cite{workman2015wide}, CVACT~\cite{liu2019lending}, and VIGOR~\cite{zhu2021vigor}, provide satellite and street-view image pairs but no textured mesh ground truth.
The closest existing resource is the VIGOR-OOD evaluation set used by Sat3DGen, which augments the Seattle test split of VIGOR with DSMs, i.e., rasterized Digital Surface Models that encode surface height.
However, it still does not provide a corresponding textured mesh asset for each satellite crop.
Sat2City v1 provides 3D supervision, but it is synthetic and uses height-map observations rather than native satellite imagery.
More broadly, we are not aware of a large-scale public dataset pairing satellite imagery with regional textured 3D meshes and calibrated pixel-to-surface correspondence. We therefore adopt an asset-centric geographic pairing strategy rather than assume pixel-level supervision: for each latitude-longitude bounding box, we export both the satellite overlay and the corresponding Google Earth~\cite{googleearth} textured 3D mesh covering the same region.
This construction yields geographically matched but naturally weakly aligned pairs: the two modalities share a geographic extent but lack calibrated pixel-to-surface correspondence (Fig.~\ref{fig:dataset_challenge}).

At the geometry level, Sat2City v1 maintains a task-specific latent space learned from a limited collection of synthetic cities.
The most direct extension would retain this representation family and relearn its VAE on real reconstructed city meshes.
We examine this route in Sec.~\ref{sec:prelim} through encode--decode reconstruction at the original resolution of 128 and after relearning the hierarchy at resolution 512.
In the shown examples, the original-resolution model loses fine urban detail, while the higher-resolution reconstruction contains fragmented geometry and blurred appearance.
These observations motivate separating representation learning from satellite-domain adaptation: \ours retains a latent space pretrained on broad, curated 3D assets and learns the satellite-to-geometry mapping within that fixed representation.

The appearance path poses a separate challenge.
Sat2City v1 generates appearance primarily from geometry and learned priors, whereas Sat2City v2 must condition appearance on the satellite image.
XCube~\cite{ren2024xcube} serves as the geometry backbone of Sat2City v1, while SCube~\cite{ren2024scube} provides the directly compatible extension for image-conditioned appearance generation.
SCube associates posed-image evidence with 3D voxels through calibrated camera geometry.
As shown in Fig.~\ref{fig:dataset_challenge}, however, our satellite-mesh pairs provide neither satellite-to-mesh camera parameters nor calibrated pixel-to-surface correspondence.
Its projection-based conditioning is therefore not directly applicable to these weakly aligned pairs.
The model must instead learn an asset-level association between satellite-image evidence and generated geometry, with the generated geometry serving as the structural anchor for material synthesis.

Accordingly, \ours reformulates Sat2City v1 by adapting a general-purpose image-conditioned 3D generator to satellite-to-city generation, rather than extending its task-specific, height-map-conditioned pipeline.
A pretrained native 3D latent space supplies a fixed representation for geospatial adaptation, while learned image-to-latent conditioning lets the satellite image guide geometry generation without calibrated pixel-to-surface correspondence.
The generated geometry then provides the structural condition for appearance synthesis.
This formulation demonstrates that weak crop-level geographic pairing can support the conditional generation of explicit 3D city assets.
The contributions are summarized as follows:
\begin{itemize}
    \item We propose \ours, to the best of our knowledge the first native structured-latent 3D framework for satellite-conditioned textured city mesh generation from a single satellite image.
    \item We construct a real-world dataset with 14,651 training pairs and 1,376 held-out test pairs, yielding 16,027 usable satellite-image--textured-mesh pairs in total. The training split covers 24 regions in 9 cities.
    %\item We adapt a pretrained native structured-latent 3D prior to weak geospatial supervision by fine-tuning a satellite-conditioned geometry flow and anchoring appearance synthesis on the generated shape.
    \item We establish an evaluation protocol for satellite-to-asset generation, under which \ours ranks first among the evaluated baselines on every reported metric across geometry accuracy, generative mesh quality, and satellite-to-asset feature alignment.
\end{itemize}

This work extends Sat2City v1, the ICCV 2025 conference version~\cite{Hua_2025_ICCV}, with three key upgrades summarized in Table~\ref{tab:version_compare}.
First, it increases the nominal geometry resolution from 128 to 512 per axis, replacing the task-specific sparse-voxel hierarchy with a pretrained compact structured-latent prior.
Second, it introduces satellite-conditioned appearance control: whereas v1 generates appearance from geometry and learned priors without satellite-image cues, v2 conditions both geometry and appearance on the satellite image, with the generated geometry anchoring material synthesis.
Third, it replaces synthetic training data with real satellite images and textured Google Earth meshes collected from matched geographic boxes.
These upgrades preserve the native 3D generation principle of v1 and are assessed through new real-world evaluation protocols and comparisons for geometry accuracy, mesh generation quality, and satellite-conditioned appearance.

\begin{table}[!t]
\centering
\caption{Upgrades from Sat2City v1 to v2. Geometry resolution is measured per axis; appearance control denotes satellite-image conditioning.}
\label{tab:version_compare}
\footnotesize
\begin{tabular*}{\columnwidth}{@{\extracolsep{\fill}}lcc@{}}
\toprule
Attribute & Sat2City v1 & Sat2City v2 \\
\midrule
Geometry resolution & 128 & 512 \\
Appearance control & \xmark & \cmark \\
Training dataset & Synthetic & Real \\
\bottomrule
\end{tabular*}
\end{table}

\section{Related Work}

\subsection{Satellite-Conditioned Urban Rendering and 3D Proxies}
Satellite-conditioned urban synthesis spans several representation regimes. Early cross-view methods map overhead imagery to individual street-level views~\cite{vid_img-lu2020geometry,vid_img-shi2022geometry}, while Sat2Vid~\cite{vid_img-li2021sat2vid} extends the task to panoramic video generation. Sat2Density~\cite{qian2023sat2density} and Sat2Density++~\cite{qian2026sat2densitypp} learn satellite-conditioned density or radiance fields for multi-view rendering, and Sat3DGen~\cite{qiansat3dgen} augments this proxy formulation with geometry-oriented constraints and perspective-view training.
At city scale, InfiniCity~\cite{city_gen-lin2023infinicity} and CityDreamer~\cite{city_gen-xie2024citydreamer} introduce compositional scene models, while GaussianCity~\cite{xie2025generative}, building on 3D Gaussian Splatting~\cite{kerbl20233d}, uses explicit Gaussian primitives for efficient city rendering.
EarthCrafter~\cite{liu2026earthcrafter} extends this direction to Earth-scale generation with Aerial-Earth3D and sparse-decoupled latent diffusion, representing appearance with textural 2D Gaussian splats. ABot-Earth 0.5~\cite{qian2026aboteearth} likewise demonstrates industrial-scale satellite-conditioned generation with 3D Gaussian Splatting.
Across these systems, the learned 3D representation primarily serves novel-view rendering. Density fields, radiance fields, triplanes, and Gaussians are therefore distinct from textured mesh assets that can be directly reused, inspected, and edited.

\subsection{Satellite-to-Explicit 3D Asset Generation}
Satellite-to-explicit asset generation instead treats geometry and appearance as primary outputs rather than intermediate states for view synthesis. One line assigns colors, textures, or learned features to obtained or generated geometry. Following the terminology of Sat3DGen~\cite{qiansat3dgen}, Sat2Scene~\cite{city_gen-li2024sat2scene} and Sat2City v1~\cite{Hua_2025_ICCV} are geometry-colorization approaches. Sat2Scene predicts pointwise colors and features for given or inferred point-cloud geometry and renders novel views. In its satellite-to-ground setting, elevation maps are lifted into vertical facades without a subsequent generative geometry-refinement stage; the dense point clouds required for high-quality rendering also favor block-level view synthesis over large-area asset generation.
Sat2City v1 instead generates native 3D geometry and appearance from synthetic height-map--asset pairs, but its appearance is not conditioned on satellite imagery. Sat2RealCity~\cite{kang2025sat2realcity} generates textured building entities from real satellite imagery with OSM spatial priors and semantic cues, focusing on buildings rather than the complete scene covered by a satellite crop.
Relative to these approaches, \ours learns from real satellite-image--textured-mesh pairs, generates the complete scene covered by the input crop, and conditions both geometry and mesh appearance on the satellite image.
A broader body of work generates persistent or controllable 3D scenes from images, sketches, layouts, or text~\cite{sce-chai2023persistent,sce-chen2023scenedreamer,sce-fridman2024scenescape,sce-hao2021gancraft,sce-xu2024sketch2scene,sce-yang2024scene123,sce-zhang20243d,yoon2026extend3d,engstler2025syncity}. These methods establish useful scene-generation mechanisms but do not address geo-referenced satellite-to-city asset generation.

\subsection{Native 3D Asset Priors and Structured Latent Models}
General-purpose 3D priors can be organized by inference procedure and output representation. Per-instance optimization and score-distillation methods synthesize objects or scenes from pretrained 2D priors~\cite{poole2022dreamfusion,bai2023componerf,zhou2024gala3d,cohen2023set,yuan2024dreamscape,jiang2024general,epstein2024disentangled,cheng2023progressive3d}, but require optimization for each sample. Sparse native 3D models such as XCube~\cite{ren2024xcube} instead learn voxel hierarchies and latent diffusion models for high-resolution shapes and outdoor scenes; Sat2City v1 follows this representation family.
Feed-forward image-to-3D systems, including LRM~\cite{img3d-hong2024lrm}, TripoSR~\cite{img3d-tochilkin2024triposr}, InstantMesh~\cite{img3d-xu2024instantmesh}, Hunyuan3D 2.0~\cite{img3d-zhao2025hunyuan3d2}, MeshGen~\cite{img3d-chen2025meshgen}, and SF3D~\cite{img3d-boss2024sf3d}, demonstrate the value of large-scale 3D and multiview pretraining. Complementary approaches use diffusion or latent sets~\cite{zhang2024clay,zhang20233dshape2vecset,zhang2024lagem} or autoregressive mesh generation~\cite{siddiqui2024meshgpt,chen2024meshanything,chen2024meshanything2,chen2024meshxl,tang2024edgerunner,weng2024pivotmesh,wang2024llama}.
TRELLIS~\cite{xiang2025trellis} uses a unified structured latent representation that captures geometry and appearance, whereas TRELLIS.2~\cite{xiang2025trellis2} introduces a more compact native representation with separate geometry and material generation. These advances provide reusable asset priors for downstream conditional generation and motivate studying how a pretrained native 3D prior can be adapted to geospatial inputs.

\subsection{Geospatial Datasets and Satellite--3D Supervision}
Geospatial datasets relevant to satellite-conditioned 3D generation can be grouped by supervision target. Cross-view benchmarks such as CVUSA~\cite{workman2015wide}, CVACT~\cite{liu2019lending}, and VIGOR~\cite{zhu2021vigor} associate satellite tiles with ground-level images or panoramas. Sat2Density++~\cite{qian2026sat2densitypp} and Sat3DGen~\cite{qiansat3dgen} use such pairs for rendering-oriented learning. VIGOR-OOD further associates the Seattle test split of VIGOR with public DSM data for geometric evaluation, but the DSM is a raster height product rather than a textured mesh and is not released as satellite-mesh training supervision.
Remote-sensing reconstruction datasets instead provide metric geometric products. DFC 2019/US3D~\cite{saux2019dfc} combines multi-view WorldView-3 imagery with DSM/LiDAR-style labels for reconstruction, while UrbanScene3D~\cite{lin2022urbanscene3d} and UrbanBIS~\cite{yang2023urbanbis} primarily contain UAV or aerial city data. Aerial-Earth3D~\cite{liu2026earthcrafter} and ABot-Earth 0.5~\cite{qian2026aboteearth} target large-scale Earth generation; ABot-Earth 0.5 aggregates proprietary acquisitions and curated public sources rather than releasing a single paired satellite-mesh corpus.
Generation-oriented resources form a third group. The Sat2City v1 dataset~\cite{Hua_2025_ICCV} contains synthetic height maps and point clouds sampled from artist-made Blender assets~\cite{blender} using CloudCompare~\cite{cloudcompare}. Its supplementary material evaluates transfer to real height maps using Depth Anything~\cite{yang2024depth} and GAMUS~\cite{xiong2023gamus}, but does not introduce real satellite-mesh training pairs. NuiScene~\cite{lee2025nuiscene} introduces the curated NuiScene43 training set, whereas Sat2RealCity~\cite{kang2025sat2realcity} provides building-level assets. None of these resources supplies the regional satellite-image--textured-mesh pairing considered here. Table~\ref{tab:dataset_compare} summarizes these supervision differences.

\begin{table}[!t]
\centering
\caption{Dataset-level comparison of satellite-related urban 3D data.}
\label{tab:dataset_compare}
\scriptsize
\setlength{\tabcolsep}{2.6pt}
\renewcommand{\arraystretch}{0.98}
\begin{tabular}{@{}p{0.52\columnwidth}c c c@{}}
\toprule
Dataset & Source & \shortstack{Satellite\\image} & \shortstack{Satellite-\\mesh pair} \\
\midrule
CVUSA~\cite{workman2015wide} & Real & \cmark & \xmark \\
CVACT~\cite{liu2019lending} & Real & \cmark & \xmark \\
VIGOR~\cite{zhu2021vigor} & Real & \cmark & \xmark \\
VIGOR-OOD~\cite{qiansat3dgen} & Real & \cmark & \xmark \\
DFC 2019 / US3D~\cite{saux2019dfc} & Real & \cmark & \xmark \\
Aerial-Earth3D~\cite{liu2026earthcrafter} & Real & \pmark & \xmark \\
UrbanScene3D~\cite{lin2022urbanscene3d} & Real & \xmark & \xmark \\
UrbanBIS~\cite{yang2023urbanbis} & Real & \xmark & \xmark \\
HoliCity~\cite{zhou2020holicity} & Real & \xmark & \xmark \\
OmniCity~\cite{li2023omnicity} & Real & \cmark & \xmark \\
GAMUS~\cite{xiong2023gamus} & Real & \cmark & \xmark \\
NuiScene43~\cite{lee2025nuiscene} & Synthetic & \xmark & \xmark \\
Sat2RealCity building assets~\cite{kang2025sat2realcity} & Real & \cmark & \xmark \\
Sat2City v1~\cite{Hua_2025_ICCV} & Synthetic & \xmark & \xmark \\
\textbf{Sat2City v2} & Real & \cmark & \cmark \\
\bottomrule
\end{tabular}
\vspace{0.2em}
\begin{flushleft}
\footnotesize
``Satellite-mesh pair'' denotes supervision in which a satellite crop is paired with a textured mesh asset from the corresponding geographic region. \cmark{} and \xmark{} indicate yes and no, respectively. \pmark{} means the dataset uses aerial or top-down imagery but not a standard satellite crop as the generation input.
\end{flushleft}
\end{table}

\begin{figure*}[!t]
\centering
\includegraphics[width=\textwidth]{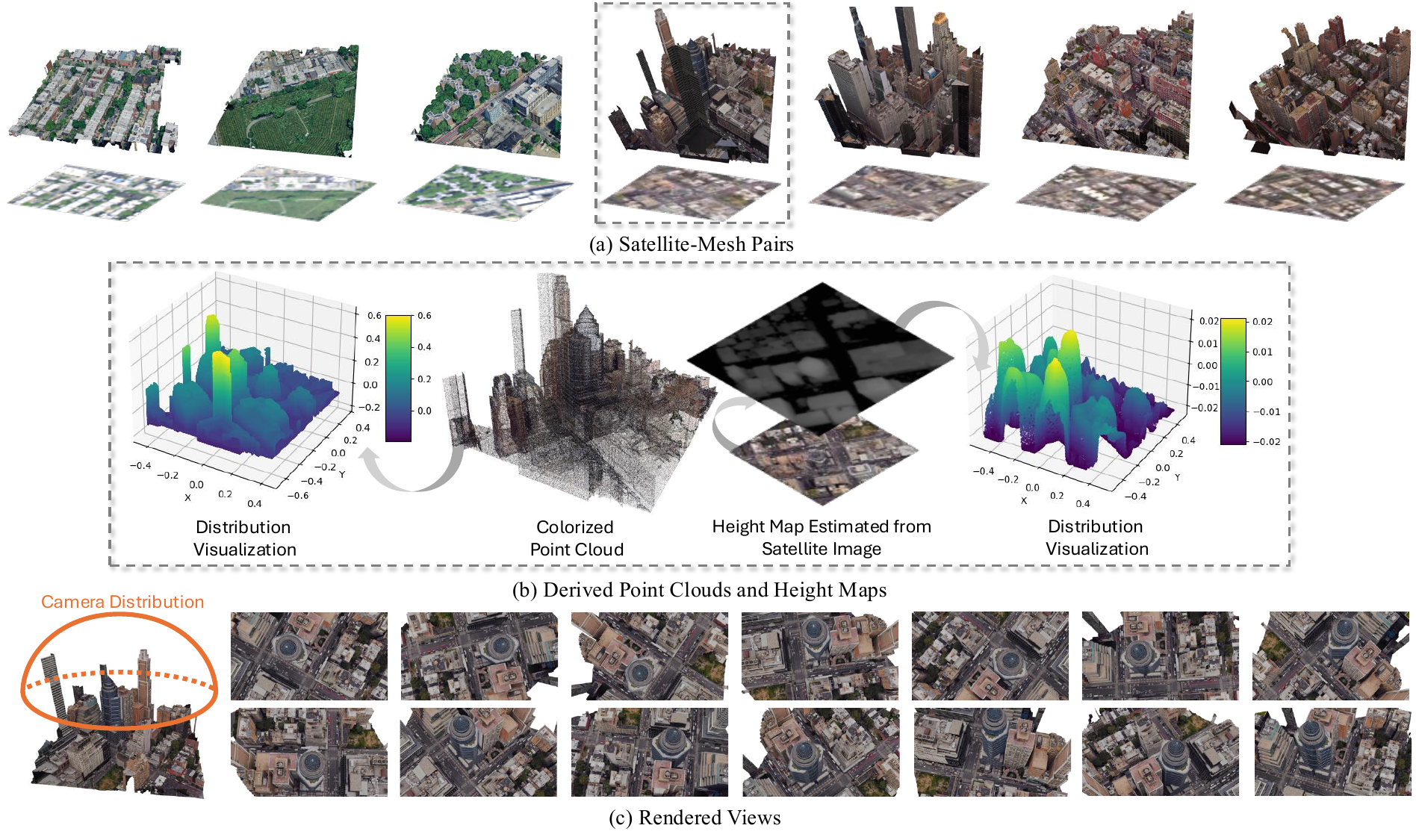}
\caption{\textbf{Overview of the Sat2City v2 dataset.} (a) \textit{Primary pair}: a geographically matched satellite image and textured 3D city mesh. (b) \textit{Auxiliary derivatives}: point clouds and learned height proxies retained for compatibility with Sat2City v1. (c) \textit{Baseline-adaptation data}: 50 rendered views with camera poses along the illustrated hemispherical trajectory. Panels (b) and (c) are not Sat2City v2 inputs or training targets.}
\label{fig:dataset}
\end{figure*}

\section{Sat2City v2 Dataset}
\label{sec:dataset}

\subsection{Data Definition}
Each Sat2City v2 sample is a tuple
\begin{equation}
\mathcal{D}_i = \{I_i^{sat}, M_i^{tex}, B_i\},
\label{eq:data_definition}
\end{equation}
where $I_i^{sat}$ is an orthorectified satellite image, $M_i^{tex}$ is a textured 3D city mesh, and $B_i$ is the latitude-longitude bounding box shared by both modalities.
The pair is weakly aligned: the satellite image and textured mesh describe the same geographic crop, but their exact geometry, time stamp, and reconstruction details may differ.
Such weak alignment reflects practical digital-twin data collected by different imagery and 3D-reconstruction pipelines; it provides asset-level geographic pairing rather than calibrated pixel-to-surface supervision.

\noindent\textbf{Model inputs versus auxiliary data.}
Equation~\eqref{eq:data_definition} defines the primary geographic record. Sat2City v2 uses $I_i^{sat}$ as its only external condition; during training, the geometry of $M_i^{tex}$ supplies the clean shape-latent target, while $B_i$ is pairing metadata. Inference requires only $I_i^{sat}$. The point clouds and learned height proxies in Fig.~\ref{fig:dataset}(b) are retained to support the point-cloud supervision and height-map conditioning used by Sat2City v1~\cite{Hua_2025_ICCV}, whereas the 50 rendered views and poses in Fig.~\ref{fig:dataset}(c) serve only baseline adaptation. None of these auxiliary products is a Sat2City v2 input or training target. Evaluation point sets and fixed-view renderings are generated separately from meshes.

\subsection{Collection and Processing}
The collection pipeline is based on Blender 4.2.1, Blender's Python API, and the free Blosm add-on for importing map-platform city tiles and satellite overlays.\footnote{\url{https://docs.blender.org/api/current/index.html}}\footnote{\url{https://github.com/vvoovv/blosm/wiki/Documentation}}
For each target city, we first define a regional latitude-longitude search window and partition it into non-overlapping city-block crops.
The tiling step uses a latitude-dependent conversion between kilometers and degrees, so a desired physical crop size is converted into crop-level geographic bounding boxes.
A Blender-based extraction pipeline then processes each crop independently, importing the Google 3D Tiles mesh and satellite overlay under the same coordinates, standardizing the material graph, and exporting the scene package summarized in Fig.~\ref{fig:dataset}.
Beyond the primary tuple in Eq.~\eqref{eq:data_definition}, each export package also stores the derivatives in Fig.~\ref{fig:dataset}(b) and the rendered views and poses in Fig.~\ref{fig:dataset}(c).
To derive the auxiliary height maps for Sat2City v1 in Fig.~\ref{fig:dataset}(b), we adapt a separate Depth Anything V2 model~\cite{yang2024depthanythingv2} to remote-sensing elevation estimation following the Depth2Elevation framework~\cite{hong2025depth2elevation}. Spatially aligned optical remote-sensing imagery and normalized digital surface model (nDSM) height maps from GAMUS~\cite{xiong2023gamus} supervise only this auxiliary estimator, not Sat2City v2.
The auxiliary estimator is then applied to all satellite images, yielding dense learned height proxies rather than elevations directly exported from the meshes.
In the current export format, the camera set contains 50 views with associated $4\times4$ pose matrices; these renderings and poses are used only when adapting view-supervised baselines, as discussed in Sec.~\ref{sec:experiments}.
Crops with invalid satellite overlays, missing city tiles, severely broken meshes, or fragmented reconstructions are removed from the active set.

After the validity checks above and the method-level preprocessing described in Sec.~\ref{sec:method}, the active training split contains 14,651 usable pairs from 24 geographic regions across nine cities.
New York contributes the largest share of samples, followed by Philadelphia, while the remaining cities broaden the long-tail coverage of geographic conditions and urban morphology.
Each valid sample contains a satellite overlay and a textured mesh for the same geographic crop.
In addition, 1,376 satellite-mesh pairs are used exclusively for testing and do not participate in training or model selection, yielding 16,027 train/test pairs in total.
The test set includes samples from New York and Philadelphia, the two dominant training cities, and an unseen-city subset from Atlanta, which is entirely absent from the training set and provides an out-of-distribution component of the aggregate held-out evaluation.

%\subsection{Why Geographically Matched Mesh Pairs Matter}
%Existing satellite-to-street datasets supervise appearance through ground-level images, but they do not provide a target 3D asset.
%DSM benchmarks supervise height but omit facades, roofs, texture maps, and reusable mesh topology.
%Synthetic city datasets provide clean assets but fail to expose temporal mismatch, missing geometry, photogrammetric noise, vegetation artifacts, and real satellite appearance.
%Geographically matched satellite-mesh pairs fill this gap.
%They do not solve exact correspondence, but they provide enough coarse signal to learn the relation between overhead imagery and city-scale 3D structure when paired with a pretrained 3D prior.

\section{Diagnostic Study of the Task-Specific Latent Space}
\label{sec:prelim}

We evaluate Sat2City v1 and its naive upgrade, denoted Sat2City v1.5, on real reconstructed city meshes from the Sat2City v2 dataset (Sec.~\ref{sec:dataset}).
This study examines the difficulties of transferring the previous framework from synthetic assets to our new real-data setting.
We begin with the VAEs because the downstream generators operate in their latent spaces and rely on their decoders to produce geometry and appearance.
Direct encode--decode reconstruction provides a first check of the representation on the new dataset, independently of learning satellite conditioning.
Specifically, we evaluate the original 128-resolution Sat2City v1 hierarchy and a naive v1.5 upgrade that retains the representation family but relearns the geometry and appearance autoencoders at a nominal geometry resolution of 512 on our real-city meshes.
%Fig.~\ref{fig:v15_architecture} places the autoencoding stage within this extension, while Fig.~\ref{fig:naive_failure} shows the corresponding reconstruction results.

\subsection{Geometry Reconstruction at the Sat2City v1 Resolution}
Sat2City v1~\cite{Hua_2025_ICCV} represents city geometry and appearance through a task-specific hierarchy of dense and sparse autoencoders learned from clean synthetic city assets.
Its VAE and decoder design follows the sparse-voxel and structure-prediction tradition of XCube and NKSR-style surface reconstruction~\cite{ren2024xcube,huang2023neural}.
The cascaded generators of Sat2City v1 follow latent diffusion and DDIM~\cite{rombach2022high,song2020ddim}, with the $v$-prediction parameterization and improved diffusion U-Net design~\cite{salimans2022v-para,dhariwal2021diffusion}.
Real reconstructed city meshes introduce narrow roads, irregular roofs, vegetation, and elevated structures alongside photogrammetric fragments, extending beyond the regular building masses in the synthetic training distribution.

We first encode and decode meshes from the Sat2City v2 dataset using the original Sat2City v1 hierarchy at nominal resolution 128.
In Fig.~\ref{fig:naive_failure}(a), fine structures are smoothed or merged into coarse blocks, limiting the visible detail of the reconstructed urban layout.
This observation motivates evaluating a higher-resolution hierarchy learned directly from real reconstructed city meshes.

\begin{figure}[!t]
\centering
\includegraphics[width=\columnwidth]{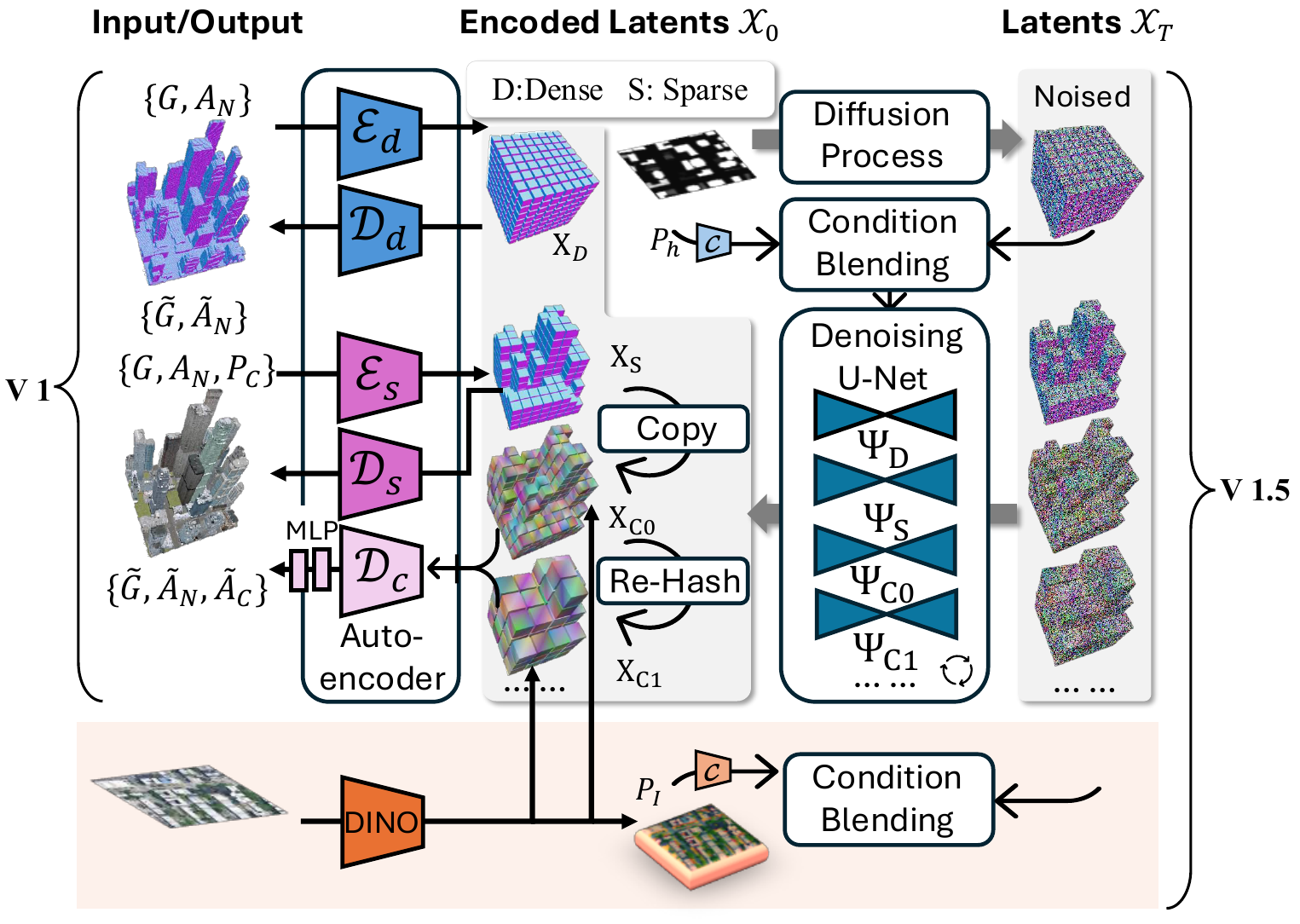}
\caption{\textbf{Architectural context for the direct v1.5 extension.} The upper block follows the Sat2City v1 pipeline~\cite{Hua_2025_ICCV}: $P_h$ is the height-map condition point cloud; $G$ is the sparse voxel geometry; $A_N$ and $A_C$ denote normal and color attributes; $P_C$ is the colorized point cloud; tildes denote decoded or generated outputs. The dense VAE encoder $\mathcal{E}_d$ and decoder $\mathcal{D}_d$ operate on the dense geometry latent $X_D$, while the sparse VAE encoder $\mathcal{E}_s$ and decoder $\mathcal{D}_s$ operate on the sparse geometry latent $X_S$. $\mathcal{D}_c$ is the appearance decoder with an MLP color head. $X_{C0}, X_{C1}, \ldots$ are multi-level appearance latents produced by Re-Hash, and $\Psi_D,\Psi_S,\Psi_{C0},\Psi_{C1},\ldots$ are the cascaded denoising U-Nets. The orange branch indicates how DINOv3 image features would enter the Sat2City v1 hierarchy in the direct v1.5 extension.}
\label{fig:v15_architecture}
\end{figure}

\subsection{Relearning the Latent Space at Higher Resolution}
For the naive Sat2City v1.5 upgrade, we increase the nominal sparse-voxel resolution from 128 to 512 while retaining the Sat2City v1 representation family, and relearn the task-specific geometry and appearance autoencoding hierarchy on the real textured city meshes of the Sat2City v2 dataset without an externally pretrained 3D asset representation.
This increases the spatial resolution while retaining dependence on the real-city meshes for learning the latent representation.
The mesh targets contain both urban structure and photogrammetric artifacts; after voxelization, both contribute to the surface occupancy supplied to the autoencoder.

\begin{figure}[!t]
\centering
\includegraphics[width=\columnwidth]{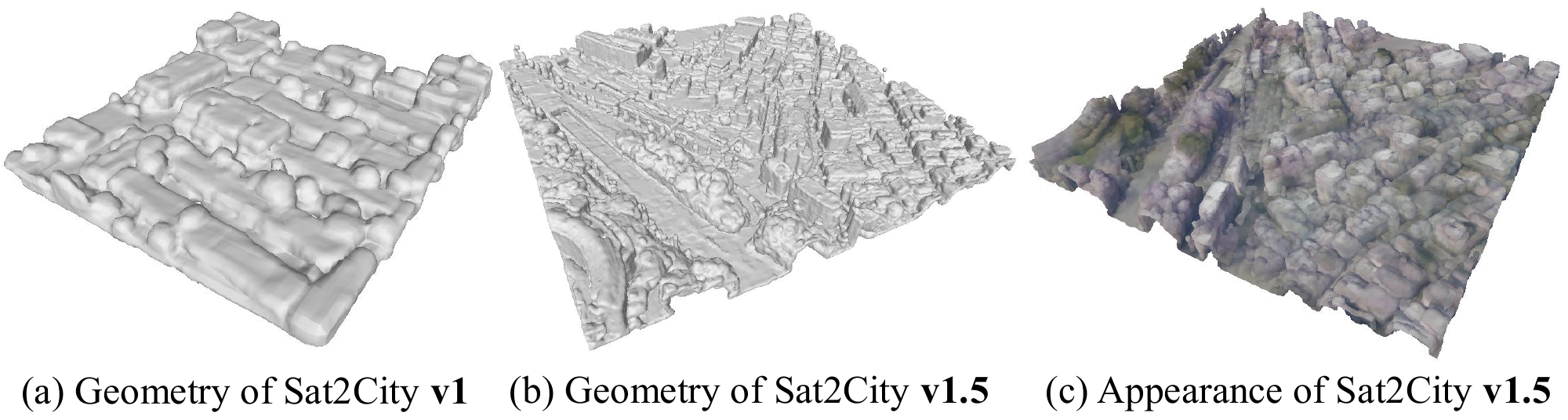}
\caption{\textbf{Encode--decode diagnostic of Sat2City v1 and its naive upgrade v1.5.} All panels show reconstructions of real meshes from the Sat2City v2 dataset. (a) The original 128-resolution Sat2City v1 hierarchy smooths fine urban structure. (b) The v1.5 hierarchy relearned at resolution 512 produces a reconstruction containing fragmented surfaces and poorly defined urban structures. (c) Its appearance reconstruction is blurred and loses fine surface detail.}
\label{fig:naive_failure}
\end{figure}

Under this training setup, the higher-resolution reconstruction in Fig.~\ref{fig:naive_failure}(b) contains fragmented surfaces and poorly defined urban structures.
The appearance reconstruction in Fig.~\ref{fig:naive_failure}(c) is likewise blurred and loses fine surface detail.
Thus, the higher-resolution retraining examined here does not yield the combination of geometric detail and coherent appearance sought for city assets.

\subsection{Implications for Sat2City v2}
The diagnostic motivates separating the need for finer spatial resolution from the choice of how to obtain the latent representation.
The original-resolution reconstruction loses fine detail, while the tested 512-resolution retraining still exhibits fragmented geometry and blurred appearance.
Because photogrammetric fragments are present in the mesh targets, their persistence alone does not establish an encoding failure; the practical concern is the visual quality of the asset reconstructions obtained in this setting.
For \ours, we therefore retain geometry and material representations pretrained on broad, curated 3D assets, avoiding the need to learn the autoencoding hierarchy solely from the reconstructed city meshes.
Within these fixed latent spaces, we adapt the satellite-conditioned geometry flow and retain the pretrained geometry-aware material pipeline, as detailed in Sec.~\ref{sec:method}.
This design combines higher-resolution geometry with satellite-conditioned appearance while using the real satellite--mesh pairs to learn the geometry mapping.
\section{Sat2City v2}
\label{sec:method}

\subsection{Overview}
\begin{figure}[!t]
\centering
\includegraphics[width=0.9\columnwidth]{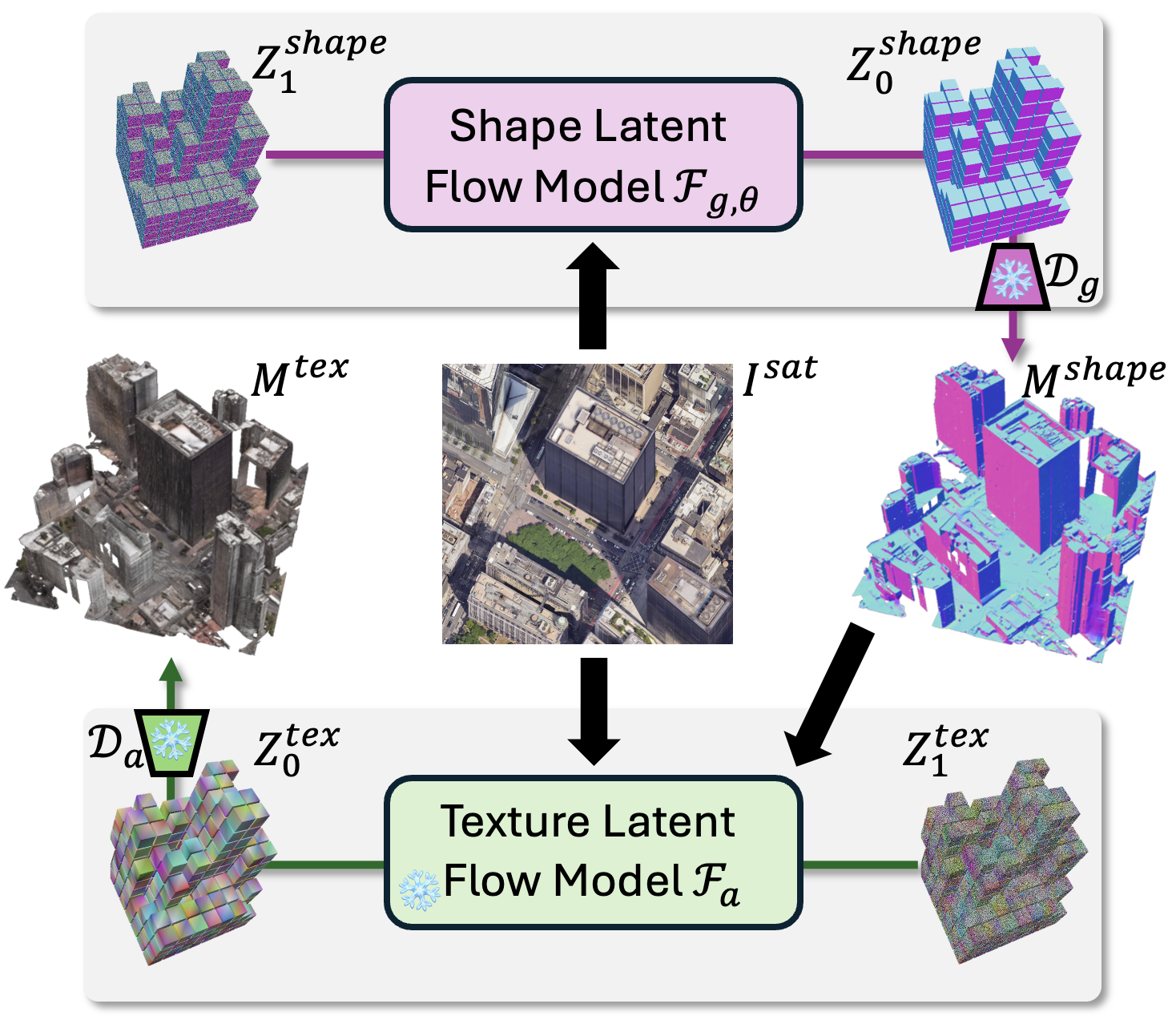}
\caption{\textbf{Pipeline of \ours.} Conditioned on $I^{sat}$, the fine-tuned geometry flow $\mathcal{F}_{g,\theta}$ transports $Z^{shape}_{1}$ to $Z^{shape}_{0}$, which the frozen $\mathcal{D}_{g}$ decodes into $M^{shape}$. The frozen texture-latent flow $\mathcal{F}_{a}$ then transports $Z^{tex}_{1}$ to $Z^{tex}_{0}$ under satellite and geometry conditions; $\mathcal{D}_{a}$ decodes material attributes that are baked onto $M^{shape}$ to produce $M^{tex}$.}
\label{fig:method}
\end{figure}

As shown in Fig.~\ref{fig:method}, \ours retains TRELLIS.2's pretrained native 3D representation and sparse-structure and material stages, adapting only its 512-resolution image-conditioned geometry flow to weakly aligned satellite--mesh pairs.
Unlike the cascaded latent diffusion of Sat2City v1, Sat2City v2 adopts conditional rectified flow matching~\cite{lipman2023flowmatching,liu2023flowstraight}.
Given a satellite image $I^{sat}$, a frozen image encoder $\Phi$ extracts conditioning tokens $C^{sat}=\Phi(I^{sat})$.
At inference, a frozen sparse-structure generator $\mathcal{S}$ first samples a binary coarse occupancy grid conditioned on $C^{sat}$.
We denote its induced conditional distribution by $p_{\mathcal{S}}$; a sample $\Omega\sim p_{\mathcal{S}}(\Omega\mid C^{sat})$ collects the coordinates of occupied cells, which define where geometry-latent tokens are placed rather than the final geometry itself.
Our satellite-conditioned geometry flow $\mathcal{F}_{g,\theta}$ then transforms Gaussian features attached to these coordinates into a terminal geometry latent $Z^{shape}_{0}$, which a frozen geometry decoder $\mathcal{D}_{g}$ converts into an explicit mesh $M^{shape}$:
\begin{equation}
\begin{aligned}
Z^{shape}_{0} &= \mathcal{F}_{g,\theta}(Z^{shape}_{1};C^{sat},\Omega),\\
M^{shape} &= \mathcal{D}_{g}(Z^{shape}_{0}).
\end{aligned}
\label{eq:geometry_pipeline}
\end{equation}
The generated mesh is subsequently processed by a frozen geometry encoder $\mathcal{E}_{g}$ and passed to a frozen geometry-aware texture-latent flow $\mathcal{F}_{a}$.
We use appearance for the rendered visual outcome; $\mathcal{F}_{a}$ specifically operates on the texture latent $Z^{tex}$, from which the appearance decoder $\mathcal{D}_{a}$ recovers PBR material attributes.
Together with features re-encoded from the same satellite image, the geometry condition guides the generation of a terminal texture latent $Z^{tex}_{0}$.
Surface baking of the decoded PBR attributes onto $M^{shape}$ is denoted by $\oplus$:
\begin{equation}
\begin{aligned}
Z^{tex}_{0} &= \mathcal{F}_{a}\!\left(Z^{tex}_{1};C^{sat},\mathcal{E}_{g}(M^{shape})\right),\\
M^{tex} &= M^{shape}\oplus\mathcal{D}_{a}(Z^{tex}_{0}).
\end{aligned}
\label{eq:appearance_pipeline}
\end{equation}
For notational compactness, we absorb the corresponding fixed inverse channel normalizations into $\mathcal{D}_{g}$ and $\mathcal{D}_{a}$.
Here, $\mathcal{F}_{g,\theta}$ and $\mathcal{F}_{a}$ denote flow maps induced by integrating their velocity fields from $t=1$ to $t=0$; $Z^{shape}_{1}$ and $Z^{tex}_{1}$ are Gaussian initial states, and $Z^{shape}_{0}$ and $Z^{tex}_{0}$ are the resulting terminal states.
Because the adopted flow path retains a residual noise scale $\sigma_{\min}$ at $t=0$, we distinguish the terminal geometry state from its exact clean training target below.

\subsection{Pretrained Native 3D Asset Prior}
Motivated by the representation diagnostics in Sec.~\ref{sec:prelim}, we inherit the native asset representation pretrained by TRELLIS.2~\cite{xiang2025trellis2} rather than learning a city-specific latent space from weakly aligned satellite-mesh pairs.
TRELLIS.2 trained its shape and material Sparse Compression VAEs on approximately 473K and 355K curated assets, respectively, and its generative prior on approximately 800K filtered assets from Objaverse-XL, ABO, HSSD, and TexVerse~\cite{deitke2023objaversexl,collins2022abo,khanna2023hssd,zhang2025texverse}.
This asset-scale pretraining covers diverse topology, fine geometry, and physically based materials before satellite-domain adaptation.
Within this representation, O-Voxels encode local dual-grid geometry and surface attributes over sparse active cells, while separate structured latent spaces decouple shape from material generation.
We therefore keep the pretrained geometry and material VAEs frozen rather than allowing noisy photogrammetric city meshes to redefine their latent manifolds.
Accordingly, $\Phi$, $\mathcal{S}$, $\mathcal{E}_{g}$, $\mathcal{D}_{g}$, $\mathcal{F}_{a}$, and $\mathcal{D}_{a}$ are inherited and frozen; only $\mathcal{F}_{g,\theta}$ is adapted to the satellite conditioning domain.

\subsection{Satellite-Conditioned Geometry Flow}
To preserve the pretrained 3D representation, each training mesh is encoded once as the clean data target of the geometry flow.
Specifically, we center the mesh, isotropically scale it to fit inside $[-0.5,0.5]^3$, transform it to the canonical coordinate system, and discretize it on a $512^3$ flexible dual grid.
The frozen geometry encoder $\mathcal{E}_{g}$ maps the resulting O-Voxel, supported at active coordinates $\Omega_i$, to raw structured-latent features $\widetilde Z_i^{shape}$, which are normalized using the fixed channel statistics $(\mu_g,\sigma_g)$ of the pretrained latent space to obtain the exact clean target $X_i^{shape}=(\widetilde Z_i^{shape}-\mu_g)/\sigma_g$.
We cache $(\Omega_i,X_i^{shape})$ for all training scenes, leaving the geometry VAE and its latent manifold unchanged.

For the geometry flow, a frozen DINOv3-L/16 encoder~\cite{simeoni2025dinov3} maps each $512\times512$ crop to 1,029 image tokens, each 1,024-dimensional: one class token, four register tokens, and 1,024 patch tokens on a $32\times32$ grid.
All 30 sparse-transformer blocks cross-attend to them; DINOv3 encodes image positions, whereas coordinate RoPE is applied to sparse 3D self-attention, enabling image-to-latent association without calibrated image-to-voxel projection.
For each clean target $X_i^{shape}$, we sample Gaussian features $\epsilon_i\sim\mathcal{N}(0,I)$ on the same sparse support $\Omega_i$.
Following conditional flow matching and rectified flow~\cite{lipman2023flowmatching,liu2023flowstraight,xiang2025trellis2}, we draw $t\sim\mathcal{U}(0,1)$ and construct
\begin{equation}
Z_{i,t}^{shape}=(1-t)X_i^{shape}
+\left[\sigma_{\min}+(1-\sigma_{\min})t\right]\epsilon_i,
\label{eq:shape_flow_path}
\end{equation}
where $\sigma_{\min}$ specifies the minimum noise scale.
Consequently, $Z_{i,1}^{shape}=\epsilon_i$ and $Z_{i,0}^{shape}=X_i^{shape}+\sigma_{\min}\epsilon_i$; $X_i^{shape}$ is the exact clean target, whereas $Z_{i,0}^{shape}$ is the near-clean terminal state.
The corresponding target velocity is $u_i=(1-\sigma_{\min})\epsilon_i-X_i^{shape}$.
The tensors $X_i^{shape}$, $\epsilon_i$, $Z_{i,t}^{shape}$, and $u_i$ all share the fixed sparse support $\Omega_i$.
In implementation, $\Omega_i$ is carried as coordinate metadata of each sparse tensor; we nevertheless expose it in the velocity notation below to make the support explicit.
The geometry-flow objective is the mean-squared error between this target and the satellite-conditioned velocity field $v_{\theta}^{g}$:
\begin{equation}
\mathcal{L}_{geo}=\mathbb{E}_{i,t,\epsilon_i}
\left[\left\|v_{\theta}^{g}(Z_{i,t}^{shape},t;C_i^{sat},\Omega_i)-u_i\right\|_2^2\right].
\label{eq:geometry_flow_loss}
\end{equation}
The parameters $\theta$ are initialized from the pretrained 512-resolution image-conditioned geometry flow and are the only generative parameters updated on Sat2City v2.
During inference, the frozen sparse-structure predictor first samples the active coordinate set $\Omega$ from $C^{sat}$; Euler integration of $v_{\theta}^{g}$ from $t=1$ to $t=0$ then maps Gaussian features at these coordinates from $Z^{shape}_{1}$ to $Z^{shape}_{0}$.

\subsection{Geometry-Anchored Appearance Synthesis}
At city scale, dominant geometric cues such as footprints, massing, and height layout are low-frequency and remain statistically informative under moderate geospatial misalignment.
Appearance, by contrast, comprises high-frequency, surface-specific color and material signals for which weakly aligned satellite-mesh pairs provide inconsistent pixel-to-surface supervision.
Although both the geometry flow and the texture-latent flow attend to image tokens, attention cannot recover correspondences absent from the training pairs; under such supervision, fine-tuning the texture-latent flow may fit inconsistent correspondences and degrade texture consistency.
We consequently retain the pretrained geometry-aware appearance pipeline of TRELLIS.2~\cite{xiang2025trellis2}.
The generated mesh $M^{shape}$ is converted to an O-Voxel and encoded by the frozen geometry encoder $\mathcal{E}_{g}$.
On the same active coordinates, the frozen texture-latent flow $\mathcal{F}_{a}$ transports the Gaussian texture state $Z^{tex}_{1}$ to $Z^{tex}_{0}$ under two complementary conditions: the encoded shape latent is concatenated channel-wise with the evolving texture state to preserve geometric correspondence, while the satellite tokens $C^{sat}$ enter through cross-attention to guide appearance attributes toward cues in the satellite observation.
The frozen appearance decoder $\mathcal{D}_{a}$ maps $Z^{tex}_{0}$ to sparse PBR fields comprising base color, metallic, roughness, and alpha.
These attributes are trilinearly sampled at the UV-rasterized mesh surface and baked into texture maps, yielding $M^{tex}$ without modifying the generated geometry.
Thus, Sat2City v2 uses this inherited appearance prior for geometry-anchored synthesis conditioned jointly on the satellite image and the generated shape.

\subsection{Implementation Details}
Our satellite-conditioned geometry flow contains 1.3B parameters and operates on latents derived from a $512^3$ O-Voxel grid.
We use AdamW for a 30,000-step training budget, with a learning rate of $2\times10^{-5}$, weight decay 0.01, bfloat16 mixed precision, adaptive gradient clipping, and an exponential moving average (EMA) decay of 0.9999.
We set $\sigma_{\min}=10^{-5}$ in Eq.~\eqref{eq:shape_flow_path} and drop the satellite condition with probability 0.1 for classifier-free guidance.
The per-GPU batch size is 4, divided into two microbatches for gradient accumulation, and training uses four NVIDIA A800-SXM4-80GB GPUs, matching the hardware configuration of Sat2City v1.
The 14,651-pair training split comprises scenes that pass the preprocessing filters in Sec.~\ref{sec:dataset}, including a maximum of 8,192 geometry-latent tokens per scene.
Satellite crops are used without object-centric background removal because the complete geographic footprint is part of the conditioning signal.
All Sat2City v2 predictions use the step-20,000 EMA checkpoint and one sample per input with random seed 42.
Inference uses a $32^3$ sparse-support grid, 512-resolution geometry, and frozen 1024-resolution appearance synthesis, exporting $2048^2$ texture maps.
The sparse-structure, geometry-latent, and texture-latent stages use the released TRELLIS.2 guidance-interval Euler sampler for 12 steps, with classifier-free guidance strengths of 7.5, 7.5, and 1.0, respectively; all remaining sampler settings follow the release~\cite{xiang2025trellis2}.

\section{Experiments}
\label{sec:experiments}

\subsection{Baselines}
We restrict main quantitative comparisons to methods that can consume the same satellite-image input or can be adapted to a clearly stated satellite-conditioned protocol.
The Sat2City v1 model is therefore discussed as historical motivation rather than included in the main tables, since it takes a satellite-derived height map instead of the native satellite image.
The direct v1.5 route is used for the representation analysis in Sec.~\ref{sec:prelim} rather than as a quantitative baseline.

We include a baseline when its released pipeline can be driven by a single satellite image without mandatory semantic maps, paired appearance-and-depth observations, or text prompts; its implementation and required resources are publicly available for reproducible evaluation; and the original work defines or discusses textured-mesh extraction from its learned 3D representation, making asset-level comparison technically meaningful.
Under these criteria, we select Sat2Density++~\cite{qian2026sat2densitypp} and Sat3DGen~\cite{qiansat3dgen} as recent satellite-conditioned urban generation methods, and additionally compare with TRELLIS.2~\cite{xiang2025trellis2} as a recent general single-image-to-3D asset generator.
EarthCrafter~\cite{liu2026earthcrafter} is not included because its released task formulations require semantic inputs or paired appearance-and-depth observations, while the available ABot-Earth 0.5~\cite{qian2026aboteearth} online interface\footnote{ABot-Earth 0.5 online demo: \url{https://abot-earth.amap.com/?tab=create}.} additionally relies on text guidance and its source code is not publicly available.

\begin{figure}[H]
\centering
\includegraphics[width=\columnwidth]{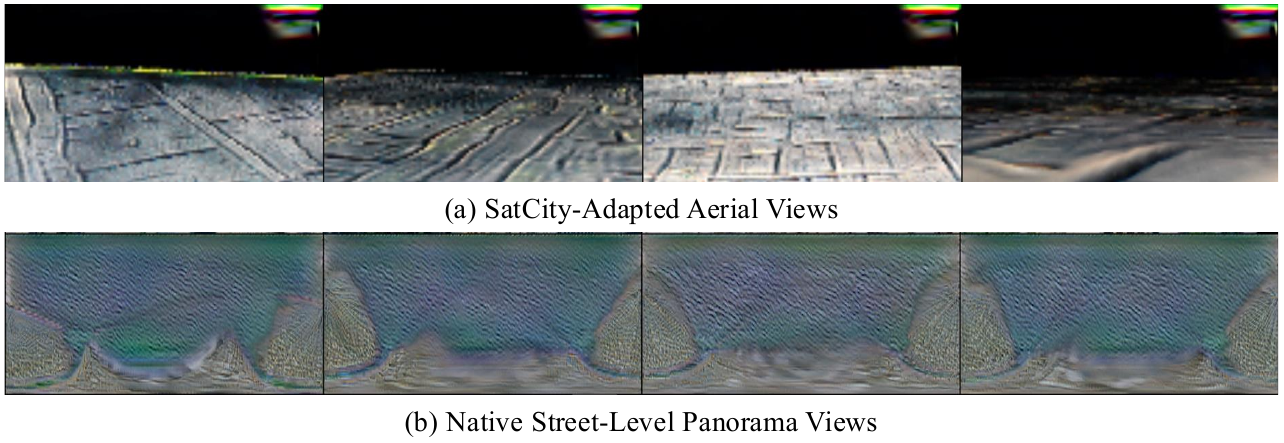}
\caption{\textbf{Direct adaptation of Sat2Density++ to Sat2City v2.} We adapt the released Sat2Density++ model~\cite{qian2026sat2densitypp} to the Sat2City v2 training data using our pinhole-camera renderings. The adapted Sat2Density++ baseline produces degraded (a) aerial views and (b) street-level views rendered from its native cylindrical $360^\circ$ panorama, consistent with the mismatch between panorama-based image formation and our pinhole-camera supervision. All panels are outputs of the adapted Sat2Density++ baseline; quantitative evaluation therefore uses its released weights.}
\label{fig:baseline_camera_mismatch}
\end{figure}

Sat2Density++ is designed around cylindrical $360^\circ$ street-view panoramas.
We directly adapt its released model to the Sat2City v2 training data using our pinhole-camera renderings, which changes both the image formation model and the distribution of supervision rays.
Under this protocol, the adapted baseline does not yield usable renderings, as shown in Fig.~\ref{fig:baseline_camera_mismatch}.
We therefore evaluate Sat2Density++ using its released weights and obtain its colored mesh through the released radiance-field mesh-export procedure.
Accordingly, the adaptation experiment is a best-effort transfer test rather than a strictly controlled comparison; the released model provides the quantitative reference.

Sat3DGen uses perspective cameras, but its non-overhead supervision comes from ground-level panoramas and perspective crops projected from them, rather than pinhole views rendered from cameras sampled along the low-altitude hemispherical trajectory in Fig.~\ref{fig:dataset}(c).
We consequently report both Sat3DGen (Original), using the released weights, and Sat3DGen (Sat2City), which retains the same architecture but is adapted on the Sat2City v2 training split using our rendered views and camera poses.
This protocol jointly changes the source data and the camera/view distribution, so Sat3DGen (Sat2City) is a best-effort transfer baseline rather than a controlled view-only or data-only ablation.
TRELLIS.2 denotes its released pretrained image-to-3D pipeline applied directly to the same satellite crops without Sat2City v2 fine-tuning.
Since \ours adapts only the geometry flow while retaining the other pretrained components, TRELLIS.2 also constitutes the ablation of \ours without satellite-domain geometry-flow fine-tuning.
This interpretation applies to all results labeled TRELLIS.2 in Tables~\ref{tab:main_geometry}--\ref{tab:appearance_semantic} and Figs.~\ref{fig:qualitative_geo} and~\ref{fig:qualitative_texture}.
All baseline outputs are subsequently processed only by the common alignment and measurement procedures specified below.

\subsection{Evaluation Protocol}
Because satellite-conditioned textured-mesh generation lacks a unified benchmark, we evaluate three complementary aspects: DSM-based geometry accuracy, generated-mesh quality, and satellite-to-asset feature alignment.

\noindent\textbf{DSM-based geometry accuracy.}
Table~\ref{tab:main_geometry} evaluates 2,882 samples from the unseen Seattle split of VIGOR~\cite{zhu2021vigor} for which both satellite imagery and LiDAR-derived ground-truth DSMs are available, following the benchmark introduced by Sat3DGen~\cite{qiansat3dgen}.
All methods receive the same satellite crop, resized to $256{\times}256$, without using street-view images, panoramas, or camera-pose information at inference time.
Sat2Density++ and Sat3DGen are evaluated through their satellite-view depth outputs, whereas for TRELLIS.2 and \ours, DSMs are obtained by orthographically rasterizing the maximum vertical coordinate of the generated meshes.
Following the released Sat3DGen evaluation implementation~\cite{qiansat3dgen}, each ground-truth DSM is spatially aligned to the corresponding prediction using a bounded planar translation and rotation search, followed by RANSAC estimation of a global vertical offset.
This alignment is used only for evaluation, and DSM ground truth is not used during training.
Because the nuisance transform is estimated with access to the reference DSM, the resulting scores measure post-registration height fidelity rather than raw georeferenced reconstruction or heading accuracy; the same procedure is applied to every method.
For valid pixels, we report MAE and RMSE by first computing each metric per image and then averaging over the test set; the fractions below 2.5 m and 7.5 m are computed over all valid pixels.

\begin{table*}[!t]
\centering
\caption{DSM-based geometry evaluation on the unseen Seattle split after the common reference-assisted registration procedure. All methods use the same satellite input and are evaluated by MAE$\downarrow$ (m), RMSE$\downarrow$ (m), $<2.5\,\mathrm{m}\uparrow$ (\%), and $<7.5\,\mathrm{m}\uparrow$ (\%). TRELLIS.2 is the \ours ablation without satellite-domain geometry-flow fine-tuning.}
\label{tab:main_geometry}
\scriptsize
\begin{tabular}{p{0.22\linewidth}cccc}
\toprule
Method & MAE$\downarrow$ (m) & RMSE$\downarrow$ (m) & $<2.5\,\mathrm{m}\uparrow$ (\%) & $<7.5\,\mathrm{m}\uparrow$ (\%) \\
\midrule
Sat2Density++ (Released)~\cite{qian2026sat2densitypp} & 4.72 & 6.76 & 49.69 & 83.65 \\
Sat3DGen (Original)~\cite{qiansat3dgen} & 3.47 & 5.20 & 62.69 & 88.68 \\
Sat3DGen (Sat2City)~\cite{qiansat3dgen} & 3.42 & 5.04 & 69.90 & 85.46 \\
TRELLIS.2~\cite{xiang2025trellis2} & 3.54 & 4.89 & 64.05 & 86.29 \\
\textbf{Sat2City v2} & \textbf{3.36} & \textbf{4.74} & \textbf{70.67} & \textbf{89.05} \\
\bottomrule
\end{tabular}
\end{table*}

\begin{table*}[!t]
\centering
\caption{Geometry generation comparison on the 1,376 held-out Sat2City v2 scenes. Following the Sat2City v1 protocol, MMD and COV use CD and EMD point-cloud distances; before their computation, all methods undergo the same reference-guided yaw search over 36 hypotheses. Normal PSNR and Normal LPIPS average normal-map similarity over four fixed $512{\times}512$ viewpoints following TRELLIS.2. The TRELLIS.2 row is the \ours ablation without satellite-domain geometry-flow fine-tuning.}
\label{tab:mesh_geometry}
\scriptsize
\begin{tabular}{p{0.16\linewidth}cccccc}
\toprule
Method & MMD (CD)$\downarrow$ & MMD (EMD)$\downarrow$ & COV (CD) \%$\uparrow$ & COV (EMD) \%$\uparrow$ & Normal PSNR$\uparrow$ & Normal LPIPS$\downarrow$ \\
\midrule
Sat2Density++ (Released)~\cite{qian2026sat2densitypp} & 0.0157 & 0.0240 & 70.64 & 45.06 & 12.29 & 0.1513 \\
Sat3DGen (Original)~\cite{qiansat3dgen} & 0.0061 & 0.0145 & 97.75 & 82.92 & 12.75 & 0.1594 \\
Sat3DGen (Sat2City)~\cite{qiansat3dgen} &0.0076 & 0.0125  & 92.65 & 83.01 & 12.07 & 0.1626 \\
TRELLIS.2~\cite{xiang2025trellis2} & 0.0045 & 0.0120 & 96.80 & 86.99 &  14.18 & 0.1186  \\
\textbf{Sat2City v2} & \textbf{0.0036} & \textbf{0.0093} & \textbf{98.36} & \textbf{89.90} & \textbf{14.53} & \textbf{0.1107}\\
\bottomrule
\end{tabular}
\end{table*}

\begin{table}[!t]
\centering
\caption{Satellite-to-asset feature alignment on the 1,376 held-out Sat2City v2 scenes. CLIP and CLIP-N use OpenAI CLIP-ViT-L/14 features and average cosine similarity across four fixed orbit views. CLIP compares satellite images with appearance renderings, whereas CLIP-N compares them with rendered normal maps. TRELLIS.2 is the \ours ablation without satellite-domain geometry-flow fine-tuning.}
\label{tab:appearance_semantic}
\scriptsize
\begin{tabular}{p{0.46\linewidth}cc}
\toprule
Method & CLIP$\uparrow$ & CLIP-N$\uparrow$ \\
\midrule
Sat2Density++ (Released)~\cite{qian2026sat2densitypp} & 0.6187 & 0.5623 \\
Sat3DGen (Original)~\cite{qiansat3dgen} & 0.7026 & 0.5980\\
Sat3DGen (Sat2City)~\cite{qiansat3dgen} & 0.7077 & 0.5788 \\
TRELLIS.2~\cite{xiang2025trellis2} & 0.7280  & 0.6414 \\
\textbf{Sat2City v2} & \textbf{0.7537} & \textbf{0.6587} \\
\bottomrule
\end{tabular}
\end{table}

\noindent\textbf{Generated-mesh geometry.}
Table~\ref{tab:mesh_geometry} and Fig.~\ref{fig:qualitative_geo} evaluate 1,376 held-out Sat2City v2 scenes, including Atlanta as a geographically out-of-distribution subset absent from training.
The reported scores aggregate the New York and Philadelphia seen-city scenes with the Atlanta unseen-city subset and therefore do not isolate out-of-distribution performance.
Each prediction and its paired reference mesh are registered in a canonical frame: the reference is centroid-translated, uniformly rescaled to the unit cube, and transformed from Blender's Z-up frame to the TRELLIS frame, while the prediction receives the centering and scale normalization associated with its native representation.
Residual global-yaw ambiguity is handled by a common reference-guided search over 36 hypotheses, selecting the rotation that minimizes Chamfer Distance (CD).
The resulting point-cloud metrics are therefore invariant to global yaw and should not be interpreted as heading accuracy; the yaw search is an evaluation operation and is not used at inference.
After simplifying each mesh to at most 100K faces, we uniformly sample 20K surface points for CD and 1,024 points for Earth Mover's Distance (EMD).
Following the Sat2City v1 protocol~\cite{Hua_2025_ICCV}, we report Minimum Matching Distance (MMD) and Coverage (COV) under both distances, with a CD-based coverage threshold of 0.02.
Visible surface detail is further assessed using the normal-map protocol of TRELLIS.2~\cite{xiang2025trellis2}: four $512{\times}512$ views are rendered at radius 10, pitch $30^\circ$, field of view (FOV) $6^\circ$, and yaw angles $30^\circ$, $120^\circ$, $210^\circ$, and $300^\circ$.
Normal PSNR and Normal LPIPS~\cite{zhang2018unreasonable} are averaged across the four views and all scenes.
\noindent\textbf{Satellite-to-asset alignment.}
Table~\ref{tab:appearance_semantic} and Fig.~\ref{fig:qualitative_texture} assess satellite-to-asset semantic correspondence on the same scenes.
Each score is referenced to the satellite image actually supplied to the corresponding model, thereby accounting for method-specific preprocessing.
For each predicted mesh, we produce appearance renderings and corresponding normal maps from four fixed orbit views at radius 2.5, pitch $45^\circ$, FOV $40^\circ$, and $512{\times}512$ resolution.
Following TRELLIS.2~\cite{xiang2025trellis2}, OpenAI CLIP-ViT-L/14~\cite{radford2021learning} encodes the satellite image and each render into L2-normalized features.
CLIP and CLIP-N are the mean view-wise cosine similarities obtained from appearance renderings and normal maps, respectively, averaged over all scenes.
CLIP-N is therefore a feature-space semantic and structural consistency proxy rather than a metric measure of 3D geometric accuracy; neither score directly measures surface-level texture fidelity to the paired reference mesh.
\subsection{Main Results}
Table~\ref{tab:main_geometry} reports DSM-based geometry accuracy.
\ours ranks first on all four metrics, achieving an MAE of 3.36 m and an RMSE of 4.74 m.
Relative to the strongest competing result in each column, it reduces RMSE from 4.89 m to 4.74 m and increases the fraction of pixels below 2.5 m error from 69.90\% to 70.67\%.
Its better point estimates than the zero-shot TRELLIS.2 ablation across both error and threshold metrics are consistent with a benefit from satellite-domain geometry-flow fine-tuning under this evaluation protocol.

\begin{figure*}[!t]
\centering
\includegraphics[width=0.90\textwidth]{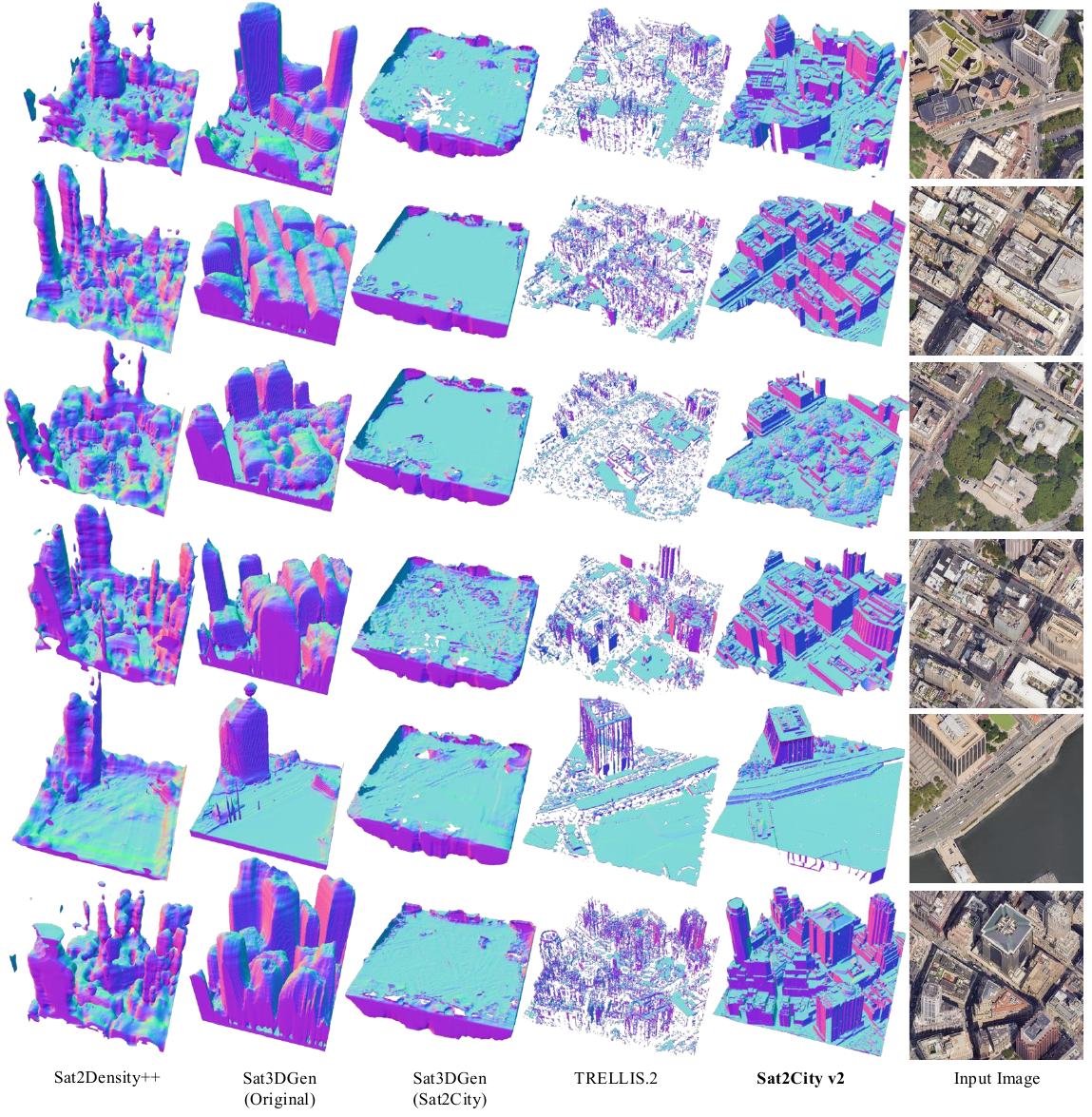}
\caption{Qualitative geometry comparison. Each row uses the same input crop and compares Sat2Density++, Sat3DGen (Original), Sat3DGen (Sat2City), TRELLIS.2, and \ours. TRELLIS.2 is the \ours ablation without satellite-domain geometry-flow fine-tuning. \ours more consistently recovers block-scale layout, building massing, and road-adjacent structure while reducing severe surface fragmentation.}
\label{fig:qualitative_geo}
\end{figure*}

Generated-mesh geometry results are reported in Table~\ref{tab:mesh_geometry} and Fig.~\ref{fig:qualitative_geo}.
\ours achieves the lowest MMD under both CD and EMD, the highest COV under both distances, the highest Normal PSNR, and the lowest Normal LPIPS.
These point estimates indicate lower distances and higher coverage under the reported protocol.
The qualitative comparison further shows improved preservation of block layout and building massing, with fewer fragmented surfaces and implausible structures than the rendering-proxy baselines.
Compared with zero-shot TRELLIS.2, \ours improves Normal PSNR from 14.18 to 14.53 and reduces Normal LPIPS from 0.1186 to 0.1107.

\begin{figure*}[!t]
\centering
\includegraphics[width=0.895\textwidth]{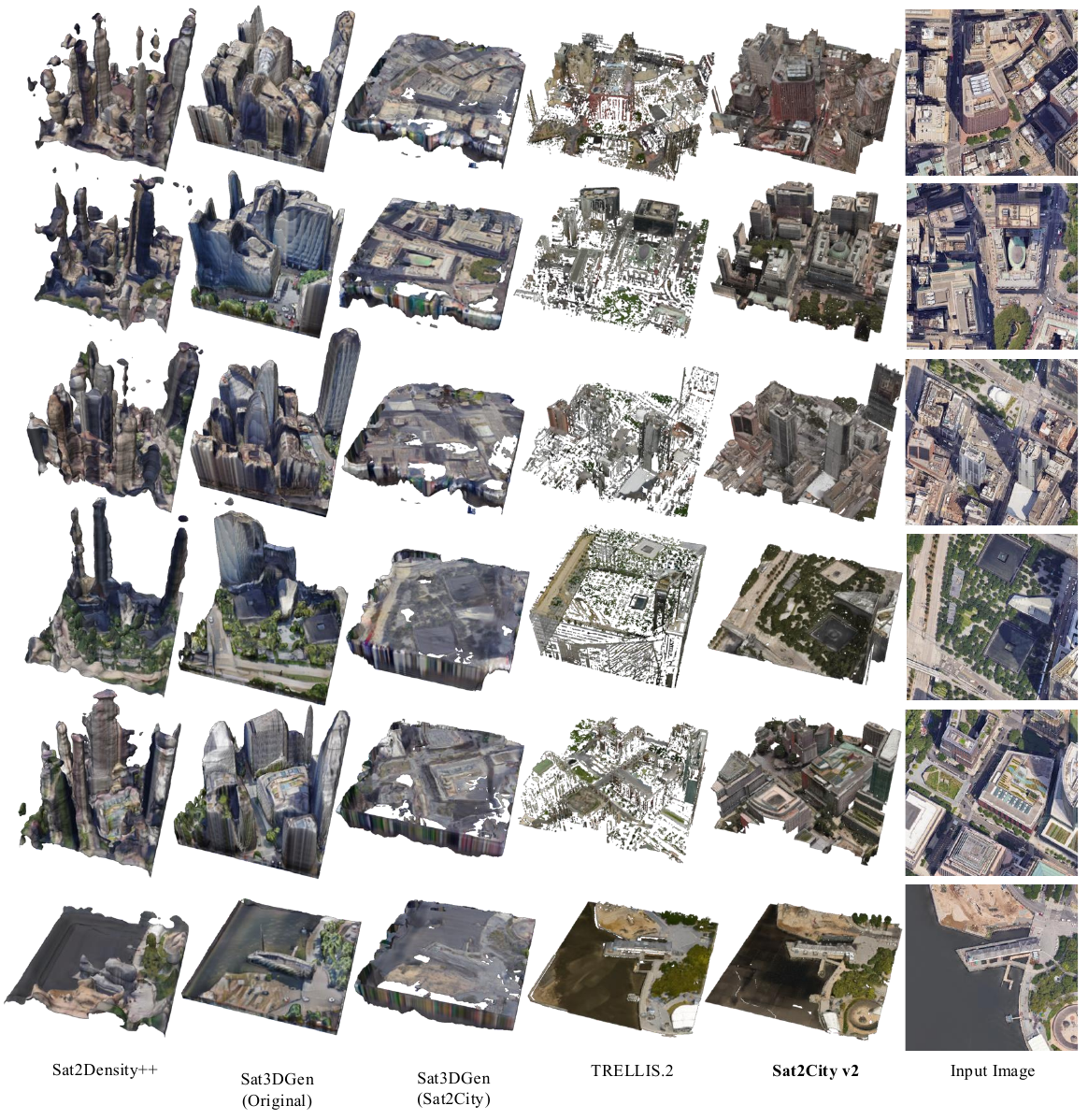}
\caption{Qualitative appearance comparison. Each row uses the same input crop and compares Sat2Density++, Sat3DGen (Original), Sat3DGen (Sat2City), TRELLIS.2, and \ours. TRELLIS.2 is the \ours ablation without satellite-domain geometry-flow fine-tuning. \ours better reflects roof, road, vegetation, and waterfront cues from the satellite input while producing coherent textured mesh assets.}
\label{fig:qualitative_texture}
\end{figure*}

Table~\ref{tab:appearance_semantic} and Fig.~\ref{fig:qualitative_texture} summarize satellite-to-asset feature alignment and qualitative appearance, respectively.
\ours attains the highest CLIP and CLIP-N scores, improving over zero-shot TRELLIS.2 from 0.7280 to 0.7537 and from 0.6414 to 0.6587, respectively.
These feature-space gains are obtained while the texture-latent flow remains frozen; they characterize the end-to-end pipeline after geometry-flow adaptation rather than a separately adapted texture model.
The qualitative results exhibit stronger correspondence in roof, road, vegetation, and ground appearance while preserving coherent textured surfaces.

\subsection{Ablation and Baseline Adaptation Analysis}
The TRELLIS.2 results in Tables~\ref{tab:main_geometry}--\ref{tab:appearance_semantic} and Figs.~\ref{fig:qualitative_geo} and~\ref{fig:qualitative_texture} constitute the ablation of \ours without satellite-domain geometry-flow fine-tuning.
In this variant, the geometry flow retains its released pretrained weights rather than learning from the satellite--mesh training pairs; the sparse-structure generator, geometry and material VAEs, and texture-latent flow remain pretrained and frozen as in \ours.
The better point estimates of \ours relative to this ablation are consistent with a benefit from adapting the image-conditioned geometry flow to geographically paired real-city data under the evaluated pipelines.
Because the texture-latent flow is frozen in both variants, the appearance feature-space differences reflect the end-to-end effect of geometry-flow adaptation without a separately adapted texture model.

The qualitative comparisons in Figs.~\ref{fig:qualitative_geo} and~\ref{fig:qualitative_texture} suggest that, in the examples shown, adaptation to Sat2City v2 changes the form of Sat3DGen's reconstruction failures rather than fully resolving them.
In the shown examples, the released model produces pronounced vertical building masses that are swollen, fused, or inconsistent with the satellite layout.
After adaptation, the outputs follow the overhead roof and ground appearance more directly, but much of the vertical structure collapses into shallow, slab-like meshes.
The corresponding normal renderings are dominated by nearly horizontal surfaces, while the appearance renderings retain missing regions and stretched textures along slab boundaries.
We hypothesize that this behavior is related to the shift from ground-level panoramas and panorama-derived perspective crops in Sat3DGen to low-altitude hemispherical renderings of photogrammetric meshes, which changes both the visible surfaces and the distribution of supervision rays.
Because the adapted model still optimizes a rendering-oriented density proxy and obtains its mesh through subsequent extraction, fitting the new view distribution does not necessarily preserve asset-level 3D structure.
This hypothesis remains qualitative because the adaptation changes the view and data distributions jointly.

%\subsection{Applications}
%\ours exports explicit OBJ/GLB assets and therefore supports workflows beyond view synthesis.
%We show applications in digital-twin block generation, controllable urban simulation, fly-through video rendering, mesh editing, and city-scale stitching by sliding-window inference.

\section{Discussion}

\subsection{Interpretation of Weak Geo-Pair Learning}
At first glance, weak satellite-mesh pairing appears too noisy for supervised learning.
The satellite image and mesh may disagree in time, occlusion, reconstruction quality, and exact alignment.
Our results are consistent with the hypothesis that weak supervision can be useful when the model is not required to learn a 3D prior from scratch: the pretrained geometry representation already encodes plausible local surface structure, so geometry-flow fine-tuning can focus on mapping overhead image evidence to that latent manifold, while the frozen material representation supplies appearance regularities downstream.
The reconstruction behavior in Sec.~\ref{sec:prelim} motivates retaining a pretrained representation, while the improvements over zero-shot TRELLIS.2 in Sec.~\ref{sec:experiments} provide evidence for learning satellite conditioning within that fixed representation.

\subsection{Limitations}
\ours still has several limitations.
The supervision is geographically matched but weakly aligned.
Expanding Sat2City v2 with satellite-mesh pairs from more cities and countries is likely to improve coverage and reduce distribution bias, but it cannot fully remove the ceiling imposed by asynchronous imagery, map tiling, and photogrammetric reconstruction artifacts.
The current training set is geographically imbalanced, with New York and Philadelphia contributing the largest shares, and its nine-city coverage may limit transfer to other countries and urban forms.
More accurate geospatial reconstruction would require more tightly coupled data, for example professional mapping assets in which satellite or aerial imagery, camera geometry, and 3D city reconstruction are maintained in a common coordinate system.
Such data are difficult to release at scale, which makes weak geo-pair learning a practical but inherently limited academic setting.

Our evaluation is also protocol dependent: DSM scores use reference-assisted planar and vertical registration, while point-cloud mesh scores use reference-guided global-yaw selection.
Although these procedures are applied identically to all methods, they assess geometry after registration rather than absolute geolocation or heading.
CLIP and CLIP-N assess satellite-to-render feature correspondence rather than direct texture fidelity, and the aggregate mixed-city results and one sample per input do not characterize city-specific out-of-distribution behavior or sampling variability; the reported point estimates also lack confidence intervals for between-method differences.
A standardized benchmark for satellite-conditioned textured-mesh generation remains unavailable.

Explicit native 3D generation and rendering-oriented representations have complementary failure modes.
Our textured meshes provide reusable geometry, topology, and material maps, but their view-level photorealism is still limited by the pretrained material prior and by the weak satellite-to-surface alignment.
By contrast, radiance-field and Gaussian-splatting representations~\cite{mildenhall2021nerf,kerbl20233d} can absorb high-frequency appearance from dense aerial or street-level observations, yet their geometry is often optimized indirectly through rendering losses and can be unreliable in complex large-scale scenes.
Recent mesh-to-Gaussian and mesh-guided Gaussian methods suggest a complementary direction in which generated meshes serve as editable geometric scaffolds for Gaussian or radiance layers that capture high-frequency and time-varying appearance~\cite{scolari2024thesis,guedon2024sugar,waczynska2024games,choi2024meshgs}.

\section{Conclusion}
We presented \ours for satellite-conditioned textured 3D city asset generation from a single satellite image.
The core shift is from a synthetic height-map-conditioned XCube-style framework to a real-data adaptation of a pretrained native structured-latent 3D asset prior.
Our encode--decode diagnostic shows fine-detail loss at the original Sat2City v1 resolution and fragmented geometry with blurred appearance after the evaluated higher-resolution retraining.
Motivated by these observations, \ours preserves the pretrained geometry and material representations, keeps the sparse-structure and material-synthesis stages frozen, and adapts only the satellite-conditioned geometry flow on weakly aligned pairs collected from matched geographic crops.
Together with the Sat2City v2 dataset, this framework provides a practical path toward satellite-conditioned generation of explicit textured 3D city assets.
Across the three reported evaluation protocols, including the registration-tolerant geometry comparisons, \ours obtains the best point estimate in every reported metric among the evaluated baselines.

%\section*{Acknowledgments}
%\todo{Add funding and institutional acknowledgments.}

\bibliographystyle{IEEEtran}
\bibliography{main}

@String(CVPR= {IEEE Conf. Comput. Vis. Pattern Recog.})

@String(ICCV= {Int. Conf. Comput. Vis.})

@String(TOG= {ACM Trans. Graph.})

@String(AAAI = {AAAI})

@IEEEtranBSTCTL{IEEEtranBSTCTL:nodash,
  CTLdash_repeated_names = "no"
}

@inproceedings{vid_img-li2021sat2vid,
  title={Sat2vid: Street-view panoramic video synthesis from a single satellite image},
  author={Li, Zuoyue and Li, Zhenqiang and Cui, Zhaopeng and Qin, Rongjun and Pollefeys, Marc and Oswald, Martin R},
  booktitle={Proceedings of the IEEE/CVF International Conference on Computer Vision},
  pages={12436--12445},
  year={2021}
}

@inproceedings{vid_img-lu2020geometry,
  title={Geometry-aware satellite-to-ground image synthesis for urban areas},
  author={Lu, Xiaohu and Li, Zuoyue and Cui, Zhaopeng and Oswald, Martin R and Pollefeys, Marc and Qin, Rongjun},
  booktitle={Proceedings of the IEEE/CVF Conference on Computer Vision and Pattern Recognition},
  pages={859--867},
  year={2020}
}

@article{vid_img-shi2022geometry,
  title={Geometry-guided street-view panorama synthesis from satellite imagery},
  author={Shi, Yujiao and Campbell, Dylan and Yu, Xin and Li, Hongdong},
  journal={IEEE Transactions on Pattern Analysis and Machine Intelligence},
  volume={44},
  number={12},
  pages={10009--10022},
  year={2022},
  publisher={IEEE}
}

@inproceedings{vid_img-deng2024streetscapes,
  title={Streetscapes: Large-scale consistent street view generation using autoregressive video diffusion},
  author={Deng, Boyang and Tucker, Richard and Li, Zhengqi and Guibas, Leonidas and Snavely, Noah and Wetzstein, Gordon},
  booktitle={ACM SIGGRAPH 2024 Conference Papers},
  pages={1--11},
  year={2024}
}

@misc{vid_img-xu2024geospecificviewgeneration,
      title={Geospecific View Generation -- Geometry-Context Aware High-resolution Ground View Inference from Satellite Views}, 
      author={Ningli Xu and Rongjun Qin},
      year={2024},
      eprint={2407.08061},
      archivePrefix={arXiv},
      primaryClass={cs.CV},
      url={https://arxiv.org/abs/2407.08061}, 
}

@article{vid_img-li2024crossviewdiff,
  title={CrossViewDiff: A Cross-View Diffusion Model for Satellite-to-Street View Synthesis},
  author={Li, Weijia and He, Jun and Ye, Junyan and Zhong, Huaping and Zheng, Zhimeng and Huang, Zilong and Lin, Dahua and He, Conghui},
  journal={arXiv preprint arXiv:2408.14765},
  year={2024}
}

@article{vid_img-li2024syntheocc,
  title={SyntheOcc: Synthesize Geometric-Controlled Street View Images through 3D Semantic MPIs},
  author={Li, Leheng and Qiu, Weichao and Cai, Yingjie and Yan, Xu and Lian, Qing and Liu, Bingbing and Chen, Ying-Cong},
  journal={arXiv preprint arXiv:2410.00337},
  year={2024}
}

@inproceedings{vid_img-yang2023urbangiraffe,
  title={Urbangiraffe: Representing urban scenes as compositional generative neural feature fields},
  author={Yang, Yuanbo and Yang, Yifei and Guo, Hanlei and Xiong, Rong and Wang, Yue and Liao, Yiyi},
  booktitle={Proceedings of the IEEE/CVF International Conference on Computer Vision},
  pages={9199--9210},
  year={2023}
}

@article{vid_img-lu2024urban,
  title={Urban Architect: Steerable 3D Urban Scene Generation with Layout Prior},
  author={Lu, Fan and Lin, Kwan-Yee and Xu, Yan and Li, Hongsheng and Chen, Guang and Jiang, Changjun},
  journal={arXiv preprint arXiv:2404.06780},
  year={2024}
}

@inproceedings{sce-hao2021gancraft,
  title={Gancraft: Unsupervised 3d neural rendering of minecraft worlds},
  author={Hao, Zekun and Mallya, Arun and Belongie, Serge and Liu, Ming-Yu},
  booktitle={Proceedings of the IEEE/CVF International Conference on Computer Vision},
  pages={14072--14082},
  year={2021}
}

@inproceedings{sce-zhang20243d,
  title={3D-SceneDreamer: Text-Driven 3D-Consistent Scene Generation},
  author={Zhang, Songchun and Zhang, Yibo and Zheng, Quan and Ma, Rui and Hua, Wei and Bao, Hujun and Xu, Weiwei and Zou, Changqing},
  booktitle={Proceedings of the IEEE/CVF Conference on Computer Vision and Pattern Recognition},
  pages={10170--10180},
  year={2024}
}

@article{sce-chen2023scenedreamer,
  title={Scenedreamer: Unbounded 3d scene generation from 2d image collections},
  author={Chen, Zhaoxi and Wang, Guangcong and Liu, Ziwei},
  journal={IEEE transactions on pattern analysis and machine intelligence},
  year={2023},
  publisher={IEEE}
}

@inproceedings{sce-chai2023persistent,
  title={Persistent nature: A generative model of unbounded 3d worlds},
  author={Chai, Lucy and Tucker, Richard and Li, Zhengqi and Isola, Phillip and Snavely, Noah},
  booktitle={Proceedings of the IEEE/CVF conference on computer vision and pattern recognition},
  pages={20863--20874},
  year={2023}
}

@inproceedings{sce-fridman2024scenescape,
  title={{SceneScape}: Text-driven consistent scene generation},
  author={Fridman, Rafail and Abecasis, Amit and Kasten, Yoni and Dekel, Tali},
  booktitle={Advances in Neural Information Processing Systems},
  volume={36},
  pages={39897--39914},
  year={2023},
  doi={10.52202/075280-1734}
}

@article{sce-yang2024scene123,
  title={Scene123: One Prompt to 3D Scene Generation via Video-Assisted and Consistency-Enhanced MAE},
  author={Yang, Yiying and Yin, Fukun and Fan, Jiayuan and Chen, Xin and Li, Wanzhang and Yu, Gang},
  journal={arXiv preprint arXiv:2408.05477},
  year={2024}
}

@article{sce-xu2024sketch2scene,
  title={Sketch2Scene: Automatic Generation of Interactive 3D Game Scenes from User's Casual Sketches},
  author={Xu, Yongzhi and Ng, Yonhon and Wang, Yifu and Sa, Inkyu and Duan, Yunfei and Li, Yang and Ji, Pan and Li, Hongdong},
  journal={arXiv preprint arXiv:2408.04567},
  year={2024}
}

@inproceedings{yoon2026extend3d,
  title={Extend3D: Town-Scale 3D Generation},
  author={Yoon, Seungwoo and Kim, Jinmo and Park, Jaesik},
  booktitle={Proceedings of the IEEE/CVF Conference on Computer Vision and Pattern Recognition},
  pages={5892--5901},
  year={2026}
}

@article{city_agent-shang2024urbanworld,
  title={UrbanWorld: An Urban World Model for 3D City Generation},
  author={Shang, Yu and Chen, Jiansheng and Fan, Hangyu and Ding, Jingtao and Feng, Jie and Li, Yong},
  journal={arXiv preprint arXiv:2407.11965},
  year={2024}
}

@article{city_agent-zhang2024cityx,
  title={CityX: Controllable Procedural Content Generation for Unbounded 3D Cities},
  author={Zhang, Shougao and Zhou, Mengqi and Wang, Yuxi and Luo, Chuanchen and Wang, Rongyu and Li, Yiwei and Yin, Xucheng and Zhang, Zhaoxiang and Peng, Junran},
  journal={arXiv preprint arXiv:2407.17572},
  year={2024}
}

@inproceedings{city_agent-yang2024procedural,
  title={Procedural Generation of 3D Scenes for Urban Landscape Based on Remote Sensing Images},
  author={Yang, Shuqin and Yuan, Haopu and Wang, Tianqi and Zhong, Rui and Song, Chenggang and Fu, Ying and Ge, Wenyi and Yuan, Xia},
  booktitle={2024 IEEE International Conference on Advanced Video and Signal Based Surveillance (AVSS)},
  pages={1--7},
  year={2024},
  organization={IEEE}
}

@inproceedings{city_gen-lin2023infinicity,
  title={Infinicity: Infinite-scale city synthesis},
  author={Lin, Chieh Hubert and Lee, Hsin-Ying and Menapace, Willi and Chai, Menglei and Siarohin, Aliaksandr and Yang, Ming-Hsuan and Tulyakov, Sergey},
  booktitle={Proceedings of the IEEE/CVF International Conference on Computer Vision},
  pages={22808--22818},
  year={2023}
}

@inproceedings{city_gen-xie2024citydreamer,
  title={Citydreamer: Compositional generative model of unbounded 3d cities},
  author={Xie, Haozhe and Chen, Zhaoxi and Hong, Fangzhou and Liu, Ziwei},
  booktitle={Proceedings of the IEEE/CVF Conference on Computer Vision and Pattern Recognition},
  pages={9666--9675},
  year={2024}
}

@article{qian2026sat2densitypp,
  title={Seeing through Satellite Images at Street Views},
  author={Qian, Ming and Tan, Bin and Wang, Qiuyu and Zheng, Xianwei and Xiong, Hanjiang and Xia, Gui-Song and Shen, Yujun and Xue, Nan},
  journal={IEEE Transactions on Pattern Analysis and Machine Intelligence},
  year={2026},
  publisher={IEEE}
}

@inproceedings{img3d-hong2024lrm,
  title={LRM: Large Reconstruction Model for Single Image to 3D},
  author={Hong, Yicong and Zhang, Kai and Gu, Jiuxiang and Bi, Sai and Zhou, Yang and Liu, Difan and Liu, Feng and Sunkavalli, Kalyan and Bui, Trung and Tan, Hao},
  booktitle={The Twelfth International Conference on Learning Representations},
  year={2024}
}

@article{img3d-tochilkin2024triposr,
  title={Triposr: Fast 3d object reconstruction from a single image},
  author={Tochilkin, Dmitry and Pankratz, David and Liu, Zexiang and Huang, Zixuan and Letts, Adam and Li, Yangguang and Liang, Ding and Laforte, Christian and Jampani, Varun and Cao, Yan-Pei},
  journal={arXiv preprint arXiv:2403.02151},
  year={2024}
}

@article{img3d-xu2024instantmesh,
  title={Instantmesh: Efficient 3d mesh generation from a single image with sparse-view large reconstruction models},
  author={Xu, Jiale and Cheng, Weihao and Gao, Yiming and Wang, Xintao and Gao, Shenghua and Shan, Ying},
  journal={arXiv preprint arXiv:2404.07191},
  year={2024}
}

@inproceedings{img3d-boss2024sf3d,
  title={Sf3d: Stable fast 3d mesh reconstruction with uv-unwrapping and illumination disentanglement},
  author={Boss, Mark and Huang, Zixuan and Vasishta, Aaryaman and Jampani, Varun},
  booktitle={Proceedings of the Computer Vision and Pattern Recognition Conference},
  pages={16240--16250},
  year={2025}
}

@article{img3d-zhao2025hunyuan3d2,
  title={Hunyuan3d 2.0: Scaling diffusion models for high resolution textured 3d assets generation},
  author={Zhao, Zibo and Lai, Zeqiang and Lin, Qingxiang and Zhao, Yunfei and Liu, Haolin and Yang, Shuhui and Feng, Yifei and Yang, Mingxin and Zhang, Sheng and Yang, Xianghui and others},
  journal={arXiv preprint arXiv:2501.12202},
  year={2025}
}

@inproceedings{img3d-chen2025meshgen,
  title={Meshgen: Generating pbr textured mesh with render-enhanced auto-encoder and generative data augmentation},
  author={Chen, Zilong and Wang, Yikai and Sun, Wenqiang and Wang, Feng and Chen, Yiwen and Liu, Huaping},
  booktitle={Proceedings of the Computer Vision and Pattern Recognition Conference},
  pages={5835--5848},
  year={2025}
}

@inproceedings{qiansat3dgen,
  title={Sat3DGen: Comprehensive Street-Level 3D Scene Generation from Single Satellite Image},
  author={Qian, Ming and Xia, Zimin and Liu, Changkun and Ma, Shuailei and Wang, Wen and Ke, Zeran and Tan, Bin and Zhang, Hang and Xia, Gui-Song},
  booktitle={The Fourteenth International Conference on Learning Representations},
  year={2026}
}

@inproceedings{liu2026earthcrafter,
  title={EarthCrafter: Scalable 3D Earth Generation via Dual-Sparse Latent Diffusion},
  author={Liu, Shang and Cao, Chenjie and Yu, Chaohui and Qian, Wen and Wang, Jing and Wang, Fan},
  booktitle={Proceedings of the AAAI Conference on Artificial Intelligence},
  volume={40},
  number={9},
  pages={7260--7268},
  year={2026},
  doi={10.1609/aaai.v40i9.37663}
}

@article{qian2026aboteearth,
  title={{ABot-Earth} 0.5: Generative 3D Earth Model},
  author={Qian, Ming and Ouyang, Tianjian and Sun, Mingchao and Wang, Zijian and Xiong, Jincheng and Han, Jiarong and Zhang, Yongchang and Zhang, Jiawei and Wang, Xu and Liu, Yu and Tang, Luyang and Yu, Fei and Ge, Zengye and Du, Mengmeng and Liu, Yuan and Fan, Nianfei and Wang, Song and Peng, Yingliang and Jia, Chunxue and Liu, Yang and Zeng, Shiying and Shi, Haozhe and Lai, Junnan and Pan, Hongyu and Wu, Zheng and Guo, Ning and Xu, Mu and Zhang, Hang},
  journal={arXiv preprint arXiv:2606.09967},
  year={2026}
}

@article{kang2025sat2realcity,
  title={Sat2RealCity: Geometry-Aware and Appearance-Controllable 3D Urban Generation from Satellite Imagery},
  author={Kang, Yijie and Wang, Xinliang and Wu, Zhenyu and Shi, Yifeng and Zhu, Hailong},
  journal={arXiv preprint arXiv:2511.11470},
  year={2025}
}

@inproceedings{xie2025generative,
  title={Generative gaussian splatting for unbounded 3D city generation},
  author={Xie, Haozhe and Chen, Zhaoxi and Hong, Fangzhou and Liu, Ziwei},
  booktitle={Proceedings of the Computer Vision and Pattern Recognition Conference},
  pages={6111--6120},
  year={2025}
}

@inproceedings{xiang2025trellis,
  title={Structured 3d latents for scalable and versatile 3d generation},
  author={Xiang, Jianfeng and Lv, Zelong and Xu, Sicheng and Deng, Yu and Wang, Ruicheng and Zhang, Bowen and Chen, Dong and Tong, Xin and Yang, Jiaolong},
  booktitle={Proceedings of the IEEE/CVF conference on computer vision and pattern recognition},
  pages={21469--21480},
  year={2025}
}

@article{xiang2025trellis2,
  title={Native and Compact Structured Latents for 3D Generation},
  author={Xiang, Jianfeng and Chen, Xiaoxue and Xu, Sicheng and Wang, Ruicheng and Lv, Zelong and Deng, Yu and Zhu, Hongyuan and Dong, Yue and Zhao, Hao and Yuan, Nicholas Jing and Yang, Jiaolong},
  journal={arXiv preprint arXiv:2512.14692},
  year={2025}
}

@InProceedings{Hua_2025_ICCV,
    author    = {Hua, Tongyan and Jiang, Lutao and Chen, Ying-Cong and Zhao, Wufan},
    title     = {Sat2City: 3D City Generation from A Single Satellite Image with Cascaded Latent Diffusion},
    booktitle = {Proceedings of the IEEE/CVF International Conference on Computer Vision (ICCV)},
    month     = {October},
    year      = {2025},
    pages     = {27978-27988}
}

@InProceedings{zhu2021vigor,
  author    = {Zhu, Sijie and Yang, Taojiannan and Chen, Chen},
  title     = {VIGOR: Cross-View Image Geo-Localization Beyond One-to-One Retrieval},
  booktitle = {Proceedings of the IEEE/CVF Conference on Computer Vision and Pattern Recognition (CVPR)},
  month     = {June},
  year      = {2021},
  pages     = {3640--3649}
}

@inproceedings{workman2015wide,
  title={Wide-Area Image Geolocalization With Aerial Reference Imagery},
  author={Workman, Scott and Souvenir, Richard and Jacobs, Nathan},
  booktitle={Proceedings of the IEEE International Conference on Computer Vision},
  pages={3961--3969},
  year={2015}
}

@inproceedings{liu2019lending,
  title={Lending Orientation to Neural Networks for Cross-View Geo-Localization},
  author={Liu, Liu and Li, Hongdong},
  booktitle={Proceedings of the IEEE/CVF Conference on Computer Vision and Pattern Recognition},
  pages={5624--5633},
  year={2019}
}

@inproceedings{qian2023sat2density,
  title={Sat2density: Faithful density learning from satellite-ground image pairs},
  author={Qian, Ming and Xiong, Jincheng and Xia, Gui-Song and Xue, Nan},
  booktitle={Proceedings of the IEEE/CVF International Conference on Computer Vision},
  pages={3683--3692},
  year={2023}
}

@inproceedings{city_gen-li2024sat2scene,
  title={Sat2Scene: 3D Urban Scene Generation from Satellite Images with Diffusion},
  author={Li, Zuoyue and Li, Zhenqiang and Cui, Zhaopeng and Pollefeys, Marc and Oswald, Martin R},
  booktitle={Proceedings of the IEEE/CVF Conference on Computer Vision and Pattern Recognition},
  pages={7141--7150},
  year={2024}
}

@article{zhang2024lagem,
  title={LaGeM: A Large Geometry Model for 3D Representation Learning and Diffusion},
  author={Zhang, Biao and Wonka, Peter},
  journal={arXiv preprint arXiv:2410.01295},
  year={2024}
}

@article{bai2023componerf,
  title={CompoNeRF: Text-guided multi-object compositional NeRF with editable 3D scene layout},
  author={Bai, Haotian and Lyu, Yuanhuiyi and Jiang, Lutao and Li, Sijia and Lu, Haonan and Lin, Xiaodong and Wang, Lin},
  journal={arXiv preprint arXiv:2303.13843},
  year={2023}
}

@article{zhou2024gala3d,
  title={Gala3d: Towards text-to-3d complex scene generation via layout-guided generative gaussian splatting},
  author={Zhou, Xiaoyu and Ran, Xingjian and Xiong, Yajiao and He, Jinlin and Lin, Zhiwei and Wang, Yongtao and Sun, Deqing and Yang, Ming-Hsuan},
  journal={arXiv preprint arXiv:2402.07207},
  year={2024}
}

@inproceedings{cohen2023set,
  title={{Set-the-Scene}: Global-local training for generating controllable {NeRF} scenes},
  author={Cohen-Bar, Dana and Richardson, Elad and Metzer, Gal and Giryes, Raja and Cohen-Or, Daniel},
  booktitle={Proceedings of the IEEE/CVF International Conference on Computer Vision Workshops},
  pages={2920--2929},
  year={2023}
}

@article{yuan2024dreamscape,
  title={DreamScape: 3D Scene Creation via Gaussian Splatting joint Correlation Modeling},
  author={Yuan, Xuening and Yang, Hongyu and Zhao, Yueming and Huang, Di},
  journal={arXiv preprint arXiv:2404.09227},
  year={2024}
}

@article{epstein2024disentangled,
  title={Disentangled 3d scene generation with layout learning},
  author={Epstein, Dave and Poole, Ben and Mildenhall, Ben and Efros, Alexei A and Holynski, Aleksander},
  journal={arXiv preprint arXiv:2402.16936},
  year={2024}
}

@article{cheng2023progressive3d,
  title={Progressive3d: Progressively local editing for text-to-3d content creation with complex semantic prompts},
  author={Cheng, Xinhua and Yang, Tianyu and Wang, Jianan and Li, Yu and Zhang, Lei and Zhang, Jian and Yuan, Li},
  journal={arXiv preprint arXiv:2310.11784},
  year={2023}
}

@inproceedings{jiang2024general,
  title={A general framework to boost 3d gs initialization for text-to-3d generation by lexical richness},
  author={Jiang, Lutao and Li, Hangyu and Wang, Lin},
  booktitle={Proceedings of the 32nd ACM International Conference on Multimedia},
  pages={6803--6812},
  year={2024}
}

@article{poole2022dreamfusion,
  title={Dreamfusion: Text-to-3d using 2d diffusion},
  author={Poole, Ben and Jain, Ajay and Barron, Jonathan T and Mildenhall, Ben},
  journal={arXiv preprint arXiv:2209.14988},
  year={2022}
}

@article{zhang2024clay,
	title={CLAY: A Controllable Large-scale Generative Model for Creating High-quality 3D Assets},
	author={Zhang, Longwen and Wang, Ziyu and Zhang, Qixuan and Qiu, Qiwei and Pang, Anqi and Jiang, Haoran and Yang, Wei and Xu, Lan and Yu, Jingyi},
	journal={ACM Transactions on Graphics (TOG)},
	volume={43},
	number={4},
	pages={1--20},
	year={2024},
	publisher={ACM New York, NY, USA}
}

@inproceedings{ren2024xcube,
	title={Xcube: Large-scale 3d generative modeling using sparse voxel hierarchies},
	author={Ren, Xuanchi and Huang, Jiahui and Zeng, Xiaohui and Museth, Ken and Fidler, Sanja and Williams, Francis},
	booktitle={Proceedings of the IEEE/CVF Conference on Computer Vision and Pattern Recognition},
	pages={4209--4219},
	year={2024}
}

@article{ren2024scube,
	title={Scube: Instant large-scale scene reconstruction using voxsplats},
	author={Ren, Xuanchi and Lu, Yifan and Liang, Hanxue and Wu, Zhangjie and Ling, Huan and Chen, Mike and Fidler, Sanja and Williams, Francis and Huang, Jiahui},
	journal={arXiv preprint arXiv:2410.20030},
	year={2024}
}

@inproceedings{siddiqui2024meshgpt,
	title={Meshgpt: Generating triangle meshes with decoder-only transformers},
	author={Siddiqui, Yawar and Alliegro, Antonio and Artemov, Alexey and Tommasi, Tatiana and Sirigatti, Daniele and Rosov, Vladislav and Dai, Angela and Nie{\ss}ner, Matthias},
	booktitle={Proceedings of the IEEE/CVF Conference on Computer Vision and Pattern Recognition},
	pages={19615--19625},
	year={2024}
}

@article{chen2024meshanything,
	title={MeshAnything: Artist-Created Mesh Generation with Autoregressive Transformers},
	author={Chen, Yiwen and He, Tong and Huang, Di and Ye, Weicai and Chen, Sijin and Tang, Jiaxiang and Chen, Xin and Cai, Zhongang and Yang, Lei and Yu, Gang and others},
	journal={arXiv preprint arXiv:2406.10163},
	year={2024}
}

@article{chen2024meshanything2,
	title={Meshanything v2: Artist-created mesh generation with adjacent mesh tokenization},
	author={Chen, Yiwen and Wang, Yikai and Luo, Yihao and Wang, Zhengyi and Chen, Zilong and Zhu, Jun and Zhang, Chi and Lin, Guosheng},
	journal={arXiv preprint arXiv:2408.02555},
	year={2024}
}

@article{chen2024meshxl,
	title={MeshXL: Neural Coordinate Field for Generative 3D Foundation Models},
	author={Chen, Sijin and Chen, Xin and Pang, Anqi and Zeng, Xianfang and Cheng, Wei and Fu, Yijun and Yin, Fukun and Wang, Yanru and Wang, Zhibin and Zhang, Chi and others},
	journal={arXiv preprint arXiv:2405.20853},
	year={2024}
}

@article{tang2024edgerunner,
	title={Edgerunner: Auto-regressive auto-encoder for artistic mesh generation},
	author={Tang, Jiaxiang and Li, Zhaoshuo and Hao, Zekun and Liu, Xian and Zeng, Gang and Liu, Ming-Yu and Zhang, Qinsheng},
	journal={arXiv preprint arXiv:2409.18114},
	year={2024}
}

@article{weng2024pivotmesh,
	title={PivotMesh: Generic 3D Mesh Generation via Pivot Vertices Guidance},
	author={Weng, Haohan and Wang, Yikai and Zhang, Tong and Chen, CL and Zhu, Jun},
	journal={arXiv preprint arXiv:2405.16890},
	year={2024}
}

@article{wang2024llama,
	title={LLaMA-Mesh: Unifying 3D Mesh Generation with Language Models},
	author={Wang, Zhengyi and Lorraine, Jonathan and Wang, Yikai and Su, Hang and Zhu, Jun and Fidler, Sanja and Zeng, Xiaohui},
	journal={arXiv preprint arXiv:2411.09595},
	year={2024}
}

@article{zhang20233dshape2vecset,
	title={3dshape2vecset: A 3d shape representation for neural fields and generative diffusion models},
	author={Zhang, Biao and Tang, Jiapeng and Niessner, Matthias and Wonka, Peter},
	journal={ACM Transactions on Graphics (TOG)},
	volume={42},
	number={4},
	pages={1--16},
	year={2023},
	publisher={ACM New York, NY, USA}
}

@inproceedings{zhang2018unreasonable,
  title={The Unreasonable Effectiveness of Deep Features as a Perceptual Metric},
  author={Zhang, Richard and Isola, Phillip and Efros, Alexei A. and Shechtman, Eli and Wang, Oliver},
  booktitle={Proceedings of the IEEE Conference on Computer Vision and Pattern Recognition},
  pages={586--595},
  year={2018}
}

@article{mildenhall2021nerf,
  title={{NeRF}: Representing scenes as neural radiance fields for view synthesis},
  author={Mildenhall, Ben and Srinivasan, Pratul P and Tancik, Matthew and Barron, Jonathan T and Ramamoorthi, Ravi and Ng, Ren},
  journal={Communications of the ACM},
  volume={65},
  number={1},
  pages={99--106},
  year={2022},
  month=jan,
  doi={10.1145/3503250},
  publisher={ACM New York, NY, USA}
}

@inproceedings{rombach2022high,
  title={High-resolution image synthesis with latent diffusion models},
  author={Rombach, Robin and Blattmann, Andreas and Lorenz, Dominik and Esser, Patrick and Ommer, Bj{\"o}rn},
  booktitle={Proceedings of the IEEE/CVF conference on computer vision and pattern recognition},
  pages={10684--10695},
  year={2022}
}

@article{lu2024infinicube,
  title={InfiniCube: Unbounded and Controllable Dynamic 3D Driving Scene Generation with World-Guided Video Models},
  author={Lu, Yifan and Ren, Xuanchi and Yang, Jiawei and Shen, Tianchang and Wu, Zhangjie and Gao, Jun and Wang, Yue and Chen, Siheng and Chen, Mike and Fidler, Sanja and others},
  journal={arXiv preprint arXiv:2412.03934},
  year={2024}
}

@misc{blender,
  author       = {Blender Foundation},
  title        = {Blender - a 3D modeling and animation software},
  year         = {2025},
  url          = {https://www.blender.org},
  note         = {Version 4.2, accessed: 2025-02-19},
}

@misc{cloudcompare,
  author       = {CloudCompare Development Team},
  title        = {CloudCompare (version 2.12.4) [GPL software]},
  year         = {2025},
  url          = {http://www.cloudcompare.org/},
  note         = {Retrieved on 2025-02-19}
}

@misc{googleearth,
  author       = {{Google}},
  title        = {Google Earth},
  year         = {2026},
  url          = {https://earth.google.com/},
  note         = {Accessed: 2026-05-28}
}

@inproceedings{huang2023neural,
  title={Neural kernel surface reconstruction},
  author={Huang, Jiahui and Gojcic, Zan and Atzmon, Matan and Litany, Or and Fidler, Sanja and Williams, Francis},
  booktitle={Proceedings of the IEEE/CVF Conference on Computer Vision and Pattern Recognition},
  pages={4369--4379},
  year={2023}
}

@article{salimans2022v-para,
  title={Progressive distillation for fast sampling of diffusion models},
  author={Salimans, Tim and Ho, Jonathan},
  journal={arXiv preprint arXiv:2202.00512},
  year={2022}
}

@article{dhariwal2021diffusion,
  title={Diffusion models beat gans on image synthesis},
  author={Dhariwal, Prafulla and Nichol, Alexander},
  journal={Advances in neural information processing systems},
  volume={34},
  pages={8780--8794},
  year={2021}
}

@article{song2020ddim,
  title={Denoising diffusion implicit models},
  author={Song, Jiaming and Meng, Chenlin and Ermon, Stefano},
  journal={arXiv preprint arXiv:2010.02502},
  year={2020}
}

@article{kerbl20233d,
  title={{3D Gaussian Splatting} for real-time radiance field rendering},
  author={Kerbl, Bernhard and Kopanas, Georgios and Leimk{\"u}hler, Thomas and Drettakis, George},
  journal={ACM Trans. Graph.},
  volume={42},
  number={4},
  pages={1--14},
  note={Art. no. 139},
  doi={10.1145/3592433},
  year={2023}
}

@mastersthesis{scolari2024thesis,
  author={Scolari, Stefano},
  title={Mesh2Splat: Gaussian Splatting from 3D Geometry and Materials},
  school={KTH Royal Institute of Technology},
  year={2024},
  url={https://urn.kb.se/resolve?urn=urn:nbn:se:kth:diva-359582}
}

@inproceedings{guedon2024sugar,
  title={SuGaR: Surface-Aligned Gaussian Splatting for Efficient 3D Mesh Reconstruction and High-Quality Mesh Rendering},
  author={Gu{\'e}don, Antoine and Lepetit, Vincent},
  booktitle={Proceedings of the IEEE/CVF Conference on Computer Vision and Pattern Recognition},
  pages={5354--5363},
  year={2024}
}

@article{waczynska2024games,
  title={GaMeS: Mesh-Based Adapting and Modification of Gaussian Splatting},
  author={Waczy{\'n}ska, Joanna and Borycki, Piotr and Tadeja, S{\l}awomir and Tabor, Jacek and Spurek, Przemys{\l}aw},
  journal={arXiv preprint arXiv:2402.01459},
  year={2024}
}

@article{choi2024meshgs,
  title={MeshGS: Adaptive Mesh-Aligned Gaussian Splatting for High-Quality Rendering},
  author={Choi, Jaehoon and Lee, Yonghan and Lee, Hyungtae and Kwon, Heesung and Manocha, Dinesh},
  journal={arXiv preprint arXiv:2410.08941},
  year={2024}
}

@article{lee2025nuiscene,
  title={NuiScene: Exploring Efficient Generation of Unbounded Outdoor Scenes},
  author={Lee, Han-Hung and Han, Qinghong and Chang, Angel X},
  journal={arXiv:2503.16375},
  year={2025}
}

@article{engstler2025syncity,
  title={SynCity: Training-Free Generation of 3D Worlds},
  author={Engstler, Paul and Shtedritski, Aleksandar and Laina, Iro and Rupprecht, Christian and Vedaldi, Andrea},
  journal={arXiv:2503.16420},
  year={2025}
}

@inproceedings{yang2024depth,
  title={Depth anything: Unleashing the power of large-scale unlabeled data},
  author={Yang, Lihe and Kang, Bingyi and Huang, Zilong and Xu, Xiaogang and Feng, Jiashi and Zhao, Hengshuang},
  booktitle={Proceedings of the IEEE/CVF Conference on Computer Vision and Pattern Recognition},
  pages={10371--10381},
  year={2024}
}

@inproceedings{yang2024depthanythingv2,
  title={{Depth Anything V2}},
  author={Yang, Lihe and Kang, Bingyi and Huang, Zilong and Zhao, Zhen and Xu, Xiaogang and Feng, Jiashi and Zhao, Hengshuang},
  booktitle={Advances in Neural Information Processing Systems},
  volume={37},
  pages={21875--21911},
  doi={10.52202/079017-0688},
  year={2024}
}

@article{xiong2023gamus,
  title={Gamus: A geometry-aware multi-modal semantic segmentation benchmark for remote sensing data},
  author={Xiong, Zhitong and Chen, Sining and Wang, Yi and Mou, Lichao and Zhu, Xiao Xiang},
  journal={arXiv preprint arXiv:2305.14914},
  year={2023}
}

@article{hong2025depth2elevation,
  title={Depth2Elevation: Scale Modulation With Depth Anything Model for Single-View Remote Sensing Image Height Estimation},
  author={Hong, Zhongcheng and Wu, Tong and Xu, Zhiyuan and Zhao, Wufan},
  journal={IEEE Transactions on Geoscience and Remote Sensing},
  year={2025},
  doi={10.1109/TGRS.2025.3564820}
}

@inproceedings{radford2021learning,
  title={Learning transferable visual models from natural language supervision},
  author={Radford, Alec and Kim, Jong Wook and Hallacy, Chris and Ramesh, Aditya and Goh, Gabriel and Agarwal, Sandhini and Sastry, Girish and Askell, Amanda and Mishkin, Pamela and Clark, Jack and Krueger, Gretchen and Sutskever, Ilya},
  booktitle={Proceedings of the International Conference on Machine Learning},
  pages={8748--8763},
  year={2021}
}

@article{saux2019dfc,
  title={2019 {IEEE GRSS Data Fusion Contest}: Large-Scale Semantic {3D} Reconstruction},
  author={Le Saux, Bertrand and Yokoya, Naoto and H{\"a}nsch, Ronny and Brown, Myron},
  journal={IEEE Geoscience and Remote Sensing Magazine},
  volume={7},
  number={4},
  pages={33--36},
  doi={10.1109/MGRS.2019.2949679},
  year={2019}
}

@article{zhou2020holicity,
  title={HoliCity: A City-Scale Data Platform for Learning Holistic 3D Structures},
  author={Zhou, Yichao and Huang, Jingwei and Dai, Xili and Luo, Linjie and Chen, Zhili and Ma, Yi},
  journal={arXiv preprint arXiv:2008.03286},
  year={2020}
}

@inproceedings{li2023omnicity,
  title={OmniCity: Omnipotent City Understanding with Multi-level and Multi-view Images},
  author={Li, Weijia and Lai, Yawen and Xu, Linning and Xiangli, Yuanbo and Yu, Jinhua and He, Conghui and Xia, Gui-Song and Lin, Dahua},
  booktitle={Proceedings of the IEEE/CVF Conference on Computer Vision and Pattern Recognition},
  pages={17397--17407},
  year={2023}
}

@inproceedings{lin2022urbanscene3d,
  title={Capturing, Reconstructing, and Simulating: The UrbanScene3D Dataset},
  author={Lin, Li and Liu, Yifei and Hu, Yue and Yan, Xin and Xie, Kai and Huang, Hui},
  booktitle={European Conference on Computer Vision},
  year={2022}
}

@inproceedings{yang2023urbanbis,
  title={{UrbanBIS}: A Large-Scale Benchmark for Fine-Grained Urban Building Instance Segmentation},
  author={Yang, Guoqing and Xue, Fuyou and Zhang, Qi and Xie, Ke and Fu, Chi-Wing and Huang, Hui},
  booktitle={ACM SIGGRAPH 2023 Conference Proceedings},
  pages={1--11},
  note={Art. no. 16},
  doi={10.1145/3588432.3591508},
  year={2023}
}

@article{simeoni2025dinov3,
  title={{DINOv3}},
  author={Sim{\'e}oni, Oriane and Vo, Huy V. and Seitzer, Maximilian and Baldassarre, Federico and Oquab, Maxime and Jose, Cijo and Khalidov, Vasil and Szafraniec, Marc and Yi, Seungeun and Ramamonjisoa, Micha{\"e}l and Massa, Francisco and Haziza, Daniel and Wehrstedt, Luca and Wang, Jianyuan and Darcet, Timoth{\'e}e and Moutakanni, Th{\'e}o and Sentana, Leonel and Roberts, Claire and Vedaldi, Andrea and Tolan, Jamie and Brandt, John and Couprie, Camille and Mairal, Julien and J{\'e}gou, Herv{\'e} and Labatut, Patrick and Bojanowski, Piotr},
  journal={arXiv preprint arXiv:2508.10104},
  year={2025}
}

@inproceedings{lipman2023flowmatching,
  title={Flow Matching for Generative Modeling},
  author={Lipman, Yaron and Chen, Ricky T. Q. and Ben-Hamu, Heli and Nickel, Maximilian and Le, Matthew},
  booktitle={International Conference on Learning Representations},
  year={2023}
}

@inproceedings{liu2023flowstraight,
  title={Flow Straight and Fast: Learning to Generate and Transfer Data with Rectified Flow},
  author={Liu, Xingchao and Gong, Chengyue and Liu, Qiang},
  booktitle={International Conference on Learning Representations},
  year={2023}
}

@inproceedings{deitke2023objaversexl,
  title={{Objaverse-XL}: A Universe of {10M+} {3D} Objects},
  author={Deitke, Matt and Liu, Ruoshi and Wallingford, Matthew and Ngo, Huong and Michel, Oscar and Kusupati, Aditya and Fan, Alan and Laforte, Christian and Voleti, Vikram and Gadre, Samir Yitzhak and VanderBilt, Eli and Kembhavi, Aniruddha and Vondrick, Carl and Gkioxari, Georgia and Ehsani, Kiana and Schmidt, Ludwig and Farhadi, Ali},
  booktitle={Advances in Neural Information Processing Systems},
  volume={36},
  pages={35799--35813},
  doi={10.52202/075280-1554},
  year={2023}
}

@inproceedings{collins2022abo,
  title={{ABO}: Dataset and Benchmarks for Real-World 3D Object Understanding},
  author={Collins, Jasmine and Goel, Shubham and Deng, Kenan and Luthra, Achleshwar and Xu, Leon and Gundogdu, Erhan and Zhang, Xi and Vicente, Tomas F. Yago and Dideriksen, Thomas and Arora, Himanshu and Guillaumin, Matthieu and Malik, Jitendra},
  booktitle={Proceedings of the IEEE/CVF Conference on Computer Vision and Pattern Recognition},
  pages={21126--21136},
  year={2022}
}

@article{khanna2023hssd,
  title={Habitat Synthetic Scenes Dataset ({HSSD}-200): An Analysis of 3D Scene Scale and Realism Tradeoffs for ObjectGoal Navigation},
  author={Khanna, Mukul and Mao, Yongsen and Jiang, Hanxiao and Haresh, Sanjay and Shacklett, Brennan and Batra, Dhruv and Clegg, Alexander and Undersander, Eric and Chang, Angel X. and Savva, Manolis},
  journal={arXiv preprint arXiv:2306.11290},
  year={2023}
}

@article{zhang2025texverse,
  title={TexVerse: A Universe of 3D Objects with High-Resolution Textures},
  author={Zhang, Yibo and Zhang, Li and Ma, Rui and Cao, Nan},
  journal={arXiv preprint arXiv:2508.10868},
  year={2025}
}

%\appendices

% \section{Author Checklist for Filling Placeholders}
% \label{app:checklist}
% Before submission, please replace all red placeholders with final material:
% \begin{itemize}
%     \item Dataset: final sample count, city/region list, OOD split, source/license, filtering rules, failed-scene count, mesh/image resolutions, and whether raw map-platform meshes can be redistributed.
%     \item Figures: teaser, motivation failure visual, dataset overview, method pipeline, qualitative comparison, and applications.
%     \item Tables: dataset comparison verification, DSM height estimation, Sat2City v2 test-set mesh geometry, and appearance semantic consistency.
%     \item Experiments: baseline reproducibility status, cameras for the 50-view rendering protocol, external DSM dataset, metrics, hardware, training/inference time.
%     \item Acknowledgments: final funding, institutional acknowledgments, and any required data/source disclaimers.
%     \item Claims: texture-latent flow is currently pretrained/frozen, sparse-structure flow is currently pretrained/frozen, Sat2City v1/v1.5 are diagnostic motivation rather than main baselines, and exact comparison against Sat3DGen/Sat2Density++ must follow the separated full-pipeline/geometry/texture tracks.
% \end{itemize}
%
\end{document}